\documentclass[10pt,journal,compsoc]{IEEEtran}
\ifCLASSOPTIONcompsoc
  \usepackage[nocompress]{cite}
\else
  \usepackage{cite}
\fi

\usepackage{amsmath}
\usepackage{wrapfig,graphicx}
\usepackage{hyperref}
\usepackage{multirow}
\usepackage{enumitem}
\usepackage{booktabs}   
\usepackage{xcolor}   
\usepackage{colortbl} 
\usepackage{pifont}
\usepackage{adjustbox}  
\usepackage{url}
\usepackage{tabu}
\usepackage{amssymb} 
\usepackage{helvet}     
\usepackage{textcomp}   
\usepackage[T1]{fontenc} 
\usepackage[para,online,flushleft]{threeparttable}
\usepackage[most]{tcolorbox}
\usepackage{xcolor}
\usepackage{tikz}

\usepackage{array}
\newcolumntype{L}[1]{>{\raggedright\let\newline\\\arraybackslash\hspace{0pt}}m{#1}}
\newcolumntype{C}[1]{>{\centering\let\newline\\\arraybackslash\hspace{0pt}}m{#1}}
\newcolumntype{R}[1]{>{\raggedleft\let\newline\\\arraybackslash\hspace{0pt}}m{#1}}

\ifCLASSINFOpdf
\else
\fi
\begin{document}
\bstctlcite{IEEEexample:BSTcontrol}
%
\title{Visual Analytics For Machine Learning: \\A Data Perspective Survey}
%
%
%
%

\author{Junpeng~Wang,~Shixia~Liu, and~Wei Zhang
\IEEEcompsocitemizethanks{\IEEEcompsocthanksitem J. Wang and W. Zhang are with Visa Research, Palo Alto,
CA, 94306.\protect\\
E-mail: \{junpenwa, wzhan\}@visa.com}
\IEEEcompsocitemizethanks{\IEEEcompsocthanksitem S. Liu is with Tsinghua University, Beijing,
China, 100084.\protect\\
E-mail: shixia@tsinghua.edu.cn.}
\thanks{Manuscript received April xx, 20xx; revised August xx, 20xx.}}

%
%

\markboth{Journal of \LaTeX\ Class Files,~Vol.~xx, No.~x, August~20xx}%
{Shell \MakeLowercase{\textit{et al.}}: Bare Demo of IEEEtran.cls for Computer Society Journals}
%



\definecolor{maroonred}{rgb}{0.69, 0.137, 0.094}
\definecolor{royalblue}{rgb}{0.259, 0.455, 0.694}

\definecolor{darkgreen}{rgb}{0.0, 0.7, 0.0}
\newcommand{\dltclr}{\textcolor{gray}} 
\newcommand{\clrr}{\textcolor{red}} 
\newcommand{\clrg}{\textcolor{darkgreen}} 
\newcommand{\clrb}{\textcolor{blue}} 
\newcommand{\revise}{\textcolor{blue}} 

\newcommand{\firstnumber}{591} 
\newcommand{\secondnumber}{555} 
\newcommand{\thirdnumber}{180}
\newcommand{\fourthnumber}{143} 
\newcommand{\dlnumber}{81} 
\newcommand{\mlnumber}{62}

\definecolor{colorinterpret}{rgb}{0.984313725,0.945098039,0.843137255}
\newtcbox{\boxinterpret}{on line,
  colframe=white,        
  colback=colorinterpret,
  coltext=black,        
  boxrule=0.1pt,        
  arc=2pt,              
  boxsep=0.1pt,
  left=2pt,right=2pt,top=1pt,bottom=1pt,
}

\definecolor{colordiagnose}{rgb}{0.980392157,0.929411765,0.611764706}
\newtcbox{\boxdiagnose}{on line,
  colframe=white,        
  colback=colordiagnose,
  coltext=black,        
  boxrule=0.1pt,        
  arc=2pt,              
  boxsep=0.1pt,
  left=2pt,right=2pt,top=1pt,bottom=1pt,
}

\definecolor{colorrefine}{rgb}{0.952941176,0.82745098,0.42745098}
\newtcbox{\boxrefine}{on line,
  colframe=white,        
  colback=colorrefine,
  coltext=black,        
  boxrule=0.1pt,        
  arc=2pt,              
  boxsep=0.1pt,
  left=2pt,right=2pt,top=1pt,bottom=1pt,
}

\definecolor{ronecolor}{rgb}{0.937, 0.745, 0.173}
\newtcbox{\taskbox}{on line,
  colframe=white,        
  colback=ronecolor,
  coltext=white,        
  boxrule=0.1pt,        
  arc=2pt,              
  boxsep=0.1pt,f
  left=2pt,right=2pt,top=1pt,bottom=1pt,
}

\definecolor{rtwocolor}{rgb}{0.518, 0.671, 0.314}

\definecolor{rthreecolor}{rgb}{0.447, 0.6, 0.824}

\definecolor{rfourcolor}{rgb}{0.686, 0.553, 0.765}

\newtcbox{\databox}{on line,
  colframe=white,        
  colback=rthreecolor,
  coltext=white,        
  boxrule=0.1pt,        
  arc=2pt,              
  boxsep=0.1pt,
  left=2pt,right=2pt,top=1pt,bottom=1pt,
}



\IEEEtitleabstractindextext{
\begin{abstract}
The past decade has witnessed a plethora of works that leverage the power of visualization (VIS) to interpret machine learning (ML) models. The corresponding research topic, VIS4ML, keeps growing at a fast pace. To better organize the enormous works and shed light on the developing trend of VIS4ML, we provide a systematic review of these works through this survey. 
Since data quality greatly impacts the performance of ML models, our survey focuses specifically on summarizing VIS4ML works from the \textbf{data perspective}. 
First, we categorize the common data handled by ML models into five types, explain the unique features of each type, and highlight the corresponding ML models that are good at learning from them. 
Second, from the large number of VIS4ML works, we tease out six tasks that operate on these types of data (i.e., data-centric tasks) at different stages of the ML pipeline to understand, diagnose, and refine ML models.
Lastly, by studying the distribution of \fourthnumber{} surveyed papers across the five data types, six data-centric tasks, and their intersections, we analyze the prospective research directions and envision future research trends.
\end{abstract}

}

\maketitle

\IEEEdisplaynontitleabstractindextext

%
\IEEEpeerreviewmaketitle

\IEEEraisesectionheading{\section{Introduction}}
\label{sec:introduction}
\IEEEPARstart{T}{he} recent success of machine learning (ML)~\cite{mitchell1997machine}, especially deep learning (DL)~\cite{lecun2015deep,goodfellow2016deep}, has received significant interest from researchers. ML has witnessed a general trend towards increasingly powerful models, however, often at the cost of being less and less interpretable. With growing concerns about the safety and reliability of ML models, their poor interpretability has started to prevent them from being adopted in many safety-critical applications, such as medical diagnosis~\cite{kwon2018retainvis,krause2017workflow} and autonomous driving~\cite{gou2020vatld, he2021can}.
To mitigate this problem, enormous visualization (VIS) efforts have been devoted to explainable artificial intelligence (XAI~\cite{molnar2020interpretable}) recently, e.g., perturbing data instances to probe ML models' decision boundary~\cite{wang2019deepvid, cheng2020dece}, training interpretable surrogates to mimic ML models' behavior~\cite{ming2018rulematrix, collaris2020explainexplore}, externalizing intermediate data from ML models to open the \textit{black-boxes}~\cite{kahng2017cti, liu2016towards}, etc. These works constitute a new research field, i.e., VIS4ML, and an increasing number of papers are being published every year in this booming field. This survey targets to structurally review them and shed light on their growing trend.

In the meantime, there is a rising tendency of shifting ML models' developments from model-centric to data-centric~\cite{strickland2022andrew}. 
Although we live in the era of big data, there are many quality issues rooted in the data, such as noisy labels~\cite{xiang2019interactive}, missing items~\cite{ye2019interactive}, and imbalanced data distributions~\cite{wang2022learning}. As the modeling techniques get more and more mature, it becomes increasingly obvious to ML developers that more performance gains could be achieved from the improvement of data rather than models. So, along with the fast and steady evolution of ML models, improving data quality for ML models attracts more research attention recently~\cite{strickland2022andrew}. This also echoes the famous proverb ``Garbage In, Garbage Out'', i.e., we can never get a satisfactory ML model without quality input data. The shift towards data-centric modeling from the ML field has also inspired many pioneering VIS works on inspecting and improving data quality through data curation, correction, and purification~\cite{xiang2019interactive,chen2020oodanalyzer,wang2020conceptexplorer}. To promote this emerging and prospective direction, we revisit and structurally review existing VIS4ML works from a \textbf{data perspective} to disclose what efforts have been conducted and what opportunities remain open. Such a review will help to inspire more VIS4ML ideas and drive more data-oriented innovations.

Our data-centric survey aims to systematically review the latest VIS4ML works by disclosing \textbf{what} data they have been focused on and \textbf{how} the data have been operated to interpret, diagnose, and refine ML models. It is carried out from the following three aspects. First, we identify the most common \textit{data types} processed by ML models, their unique features, and how ML models have been tailored to better learn from them (Sec. \ref{sec:data}). Second, focusing on the operations applied to the identified data types, we elicit six \textit{data-centric VIS4ML tasks} serving the general goal of model understanding, diagnosis, and refinement~\cite{choo2018visual, hohman2018visual} (Sec. \ref{sec:task}). Third, by studying the distribution of the surveyed papers across different data types, VIS4ML tasks, and their intersections, we summarize the ongoing research trend and disclose prospective VIS4ML research directions (Sec. \ref{sec:trend}).

In essence, the contributions of our survey are twofold. First, we provide a data-centric taxonomy for VIS4ML and comprehensively review the latest works following the taxonomy. The taxonomy and review help researchers better understand the fast-growing number of VIS4ML works, reexamine them in a new angle, and unblock researchers from proposing more data-centric VIS4ML works. Second, from the coverage of the surveyed papers across different taxonomy sub-categories, we reveal what data types, VIS4ML tasks, or data-task combinations have not been sufficiently explored,
pointing the way to promising research directions and nourishing new ideas in this flourishing field. 
An interactive webpage for the survey has been developed using SurVis~\cite{beck2015visual}, which is available at: \url{https://vis4ml.github.io/}.

\section{Related Works}
\label{sec:relatedsurvey}
\textbf{Existing VIS4ML Surveys.} As the number of VIS4ML works keeps growing at a fast pace, there have been multiple surveys~\cite{chatzimparmpas2020state, lu2017state, yuan2021survey,choo2018visual, hohman2018visual} and conceptual frameworks~\cite{spinner2019explainer,sacha2018vis4ml} trying to organize and review them. We discuss these works and highlight the unique perspective that we have taken to differentiate our survey from them.

\textit{Task-Centric.} Based on the tasks that VIS works try to accomplish when serving ML, researchers have categorized VIS4ML works into \texttt{understanding}, \texttt{diagnosing}, and \texttt{refining} ML models~\cite{choo2018visual, hohman2018visual, spinner2019explainer}. These three tasks have been well-recognized by the VIS community and referred to in many latest papers~\cite{liu2017towards, wang2019interpreting}. We also advocate this categorization and consider these tasks as three high-level goals of VIS4ML. Our survey further distills six low-level tasks that are often performed when accomplishing these goals (Fig.~\ref{fig:taxonomy}d). For example, \texttt{refining} a model can be achieved by \texttt{generating} new data or \texttt{improving} existing data (two low-level tasks). Moreover, although model refinement can be conducted from both the model side (e.g., architecture pruning~\cite{pezzotti2017deepeyes,li2020cnnpruner}) and data side, we limit ourselves on the data side to present unique data-level insights.

\textit{Procedure-Centric.} Following the building process of ML models and existing ML pipelines, Yuan et al.~\cite{yuan2021survey} separated VIS works into groups that interpret ML models \texttt{before}, \texttt{during}, and \texttt{after} their building process. Likewise, by tracing ML models' execution, Chatzimparmpas et al.~\cite{chatzimparmpas2020state} reviewed how VIS enhances the trust-level in five key ML pipeline stages. 
The predictive visual analytics framework followed similar stages to review VIS works for predictive models~\cite{lu2017state}. Our survey also considers the ML construction pipelines. Following our unique focus of ML data, we identify the ``operational data'' from ML pipelines as \texttt{input}, \texttt{intermediate}, and \texttt{output} data (Fig.~\ref{fig:taxonomy}b) and explain what data-centric tasks are often conducted on each of them.
    
\textit{Human/User-Centric.} By analyzing human involvements in different ML model-building stages, Sacha et al.~\cite{sacha2018vis4ml} introduced a human-centric VIS4ML ontology, where VIS assists humans to \texttt{prepare-data}, \texttt{prepare-learning}, \texttt{model-learning}, and \texttt{evaluate-model}. Similarly, there are multiple attempts trying to exploit the role of \textit{users} in exploratory model analysis~\cite{cashman2019user} and active learning~\cite{bernard2018towards}. 
This user-centric viewpoint diverges significantly from our data-centric perspective, resulting in distinct paper categorizations and unique insights from respective standpoints.

There is a great overlap of the covered papers between our survey and the earlier VIS4ML surveys. However, the reviewing perspective and paper categorizations of our survey are very different from those of the earlier ones, so as to the disclosed insights and identified research opportunities. 
For example, Yuan et al.~\cite{yuan2021survey} took a procedure-centric perspective and discussed the input data-quality issues in their \texttt{before} model-building category. In our survey, the same issue is discussed in the \texttt{assess} task (Sec.~\ref{sec:assess_input}). Due to this overlap, both surveys identify the research opportunity of improving data quality, echoing its importance. On the other hand, our \texttt{assess} task also covers the output data assessment (Sec.~\ref{sec:assess_output}), which (partially) corresponds to the \texttt{after} model-building category of Yuan et al.~\cite{yuan2021survey}. Although the studied VIS4ML papers may largely overlap, different perspectives organize them into different groups, disclosing unique insights from respective perspectives. For example, the research opportunities identified from Yuan et al.~\cite{yuan2021survey} are what can be further improved \texttt{before}/\texttt{after} model-building. In contrast, our survey will provide insights into \textit{what} data types have been under-explored and \textit{how} the data can be further \texttt{assessed}.

\textbf{Existing VIS Task Taxonomies.} As we introduce a taxonomy for data-centric VIS4ML tasks, the existing VIS task taxonomies are also related to our work. Amar et al.~\cite{amar2005low} summarized 10 low-level tasks to accomplish the high-level goal of data understanding. In analogy to their rationale, we summarize six low-level tasks to accomplish the three high-level VIS4ML goals of understanding, diagnosing, and refining ML models~\cite{choo2018visual}. 
Brehmer and Munzner~\cite{brehmer2013multi} organized VIS tasks into a multi-level topology, which answers \textit{why} a task is performed, \textit{how} it is performed, and \textit{what} the task's input and output are. As emphasized by the authors, their tasks are abstract with no target applications, so that they can compare them across applications. In contrast, our tasks here are specific to VIS4ML and they all focus on ML operational data. 
The tasks introduced by Shneiderman~\cite{shneiderman1996eyes} are designated for data exploration (e.g., \texttt{zoom}/\texttt{filter}) but are not able to cover the diverse aspects in ML model analysis, such as data assessment and data improvement. 
Given these differences, we cannot directly reuse existing taxonomies. With iterative explorations and progressive refinements (detailed in Sec.~\ref{sec:rational}), we came out with the methodology of deriving data-centric VIS4ML tasks by carefully examining the requirement/task analysis section of individual VIS4ML papers, and finally elicited six tasks (Sec.~\ref{sec:task}). Note that there are definitely overlaps between our tasks and the tasks from the existing VIS literature, as VIS4ML is a subdomain of VIS. For example, the essence of our \texttt{present} task is similar to the \texttt{present} task in~\cite{brehmer2013multi} and the \texttt{overview} task in~\cite{shneiderman1996eyes}. Nevertheless, our identified tasks are always data-centric and specific to the VIS4ML domain.

\begin{figure*}[tbh]
 \centering 
\includegraphics[width=\textwidth]{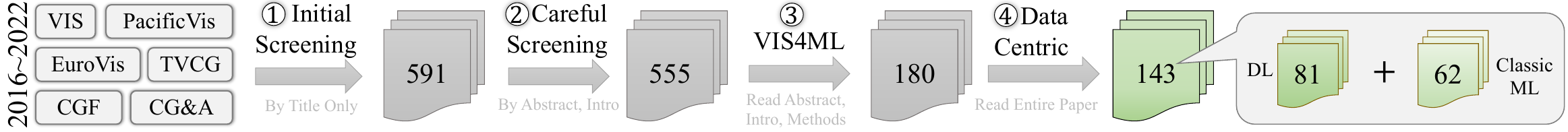}
\vspace{-0.3in}
 \caption{Four-step paper selection process. The covered \fourthnumber{} papers include \dlnumber{} DL-specific ones and \mlnumber{} ones for the interpretation of classic ML.}
 \label{fig:process}
 \vspace{-0.1in}
\end{figure*}

\section{Survey Landscape and Taxonomy}
We have seen an increasing number of VIS4ML works since 2016, and thus, set the temporal coverage of our survey to be 2016-2022.
Within this temporal range, we identify related VIS4ML works by screening research papers from major VIS conferences and journals, including:
\begin{itemize}[leftmargin=11pt, topsep=0pt,itemsep=0pt,parsep=0pt,partopsep=0pt]
\item \textit{IEEE Visualization \& Visual Analytics Conference (VIS),} 
\item \textit{Eurographics Conference on Visualization (EuroVis),}
\item \textit{IEEE Pacific Visualization Symposium (PacificVis),} 
\item \textit{IEEE Trans. on Visualization and Computer Graphics (TVCG),} 
\item \textit{Computer Graphics Forum (CGF),}  
\item \textit{Computer Graphics \& Applications (CG\&A).} 
\end{itemize}

\subsection{Paper Selection}
From the covered venues and specified temporal range, we extract the related VIS4ML papers in four steps (Fig.~\ref{fig:process}). \textbf{First}, an initial screening of all papers in the range is conducted, focusing majorly on the papers' title to decide if they are related to VIS+ML or not. 
This process filters out \firstnumber{} papers and we have included them all in our Supplementary Material. 
\textbf{Second}, for a more careful screening of the filtered papers, we read their Abstract and Introduction to exclude papers that are actually not related to ML (though their title includes some related words, such as ``Learning'' or ``Deep''). 
This step reduces the number of papers down to \secondnumber{}. \textbf{Third}, we read the methodology sections of the papers and check their included figures to exclude ML4VIS works. These works use ML to solve traditional VIS problems or facilitate data analysis, but present less model interpretation effort. A large number of papers belong to this category and excluding them reduces the number of papers down to \thirdnumber{}. \textbf{Lastly}, for the remaining papers, we further exclude works that (1) focus solely on the interpretation of ML models' architectures or hyperparameters where data is not their focus, or (2) introduce conceptual frameworks (or positional papers) that do not perform any data operations. 
For (1), Net2Vis~\cite{bauerle2021net2vis} introduces a grammar to easily extract CNN architectures and visualize them as publication-tailored figures. 
DNN Genealogy~\cite{wang2019visual} summarizes the evolution trend of DNN architectures and conducts visual analytics on the trend. Both papers present great VIS4ML contributions. However, since they focus solely on models' architecture and no data is involved, we exclude them from this data-centric survey. The papers in (2) organize and review existing VIS4ML works from different angles, e.g., ~\cite{chatzimparmpas2020state, spinner2019explainer,sacha2018vis4ml}. However, since they do not conduct concrete data operations, they have also been excluded.

\begin{figure}[tbh]
 \centering 
\includegraphics[width=\columnwidth]{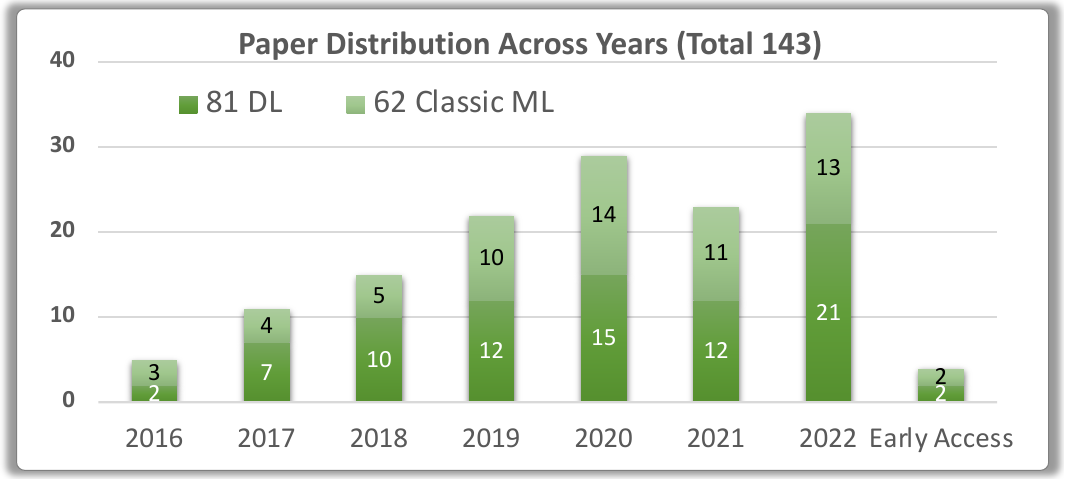}
  \vspace{-0.3in}
 \caption{The surveyed \fourthnumber{} papers (\dlnumber{}/\mlnumber{} for DL/classic ML) over years.}
 \label{fig:years}
\end{figure}

Finally, \fourthnumber{} closely related VIS4ML papers have been identified. Among them, \dlnumber{} focus specifically on the interpretation of DL models, whereas the rest \mlnumber{} interpret classic ML models (e.g., decision trees and SVMs) or their proposed solution is general enough for any ML models. The papers'  distribution across years is shown in Fig.~\ref{fig:years}. An increasing trend is clearly observed (for both \textit{DL} and \textit{classic ML}).

\begin{figure*}[tbh]
 \centering 
\includegraphics[width=\textwidth]{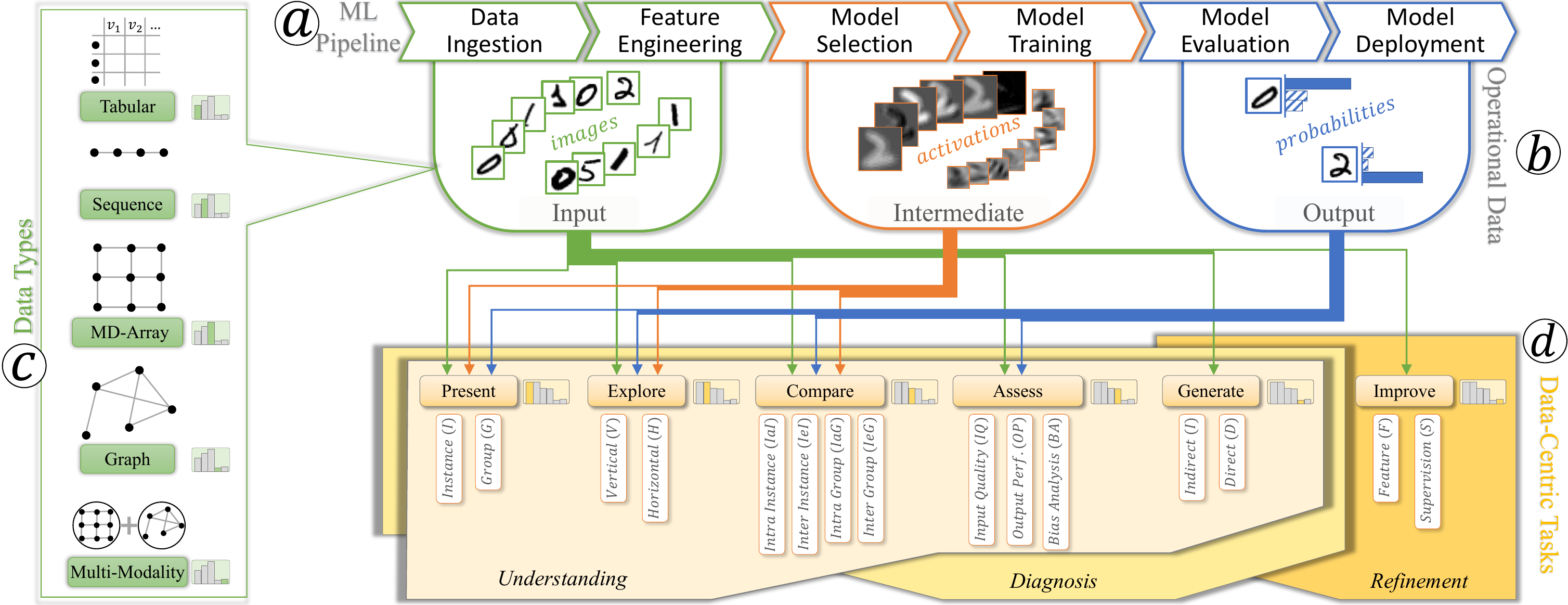}
  \vspace{-0.3in}
 \caption{The data-centric taxonomy. The \textbf{\textit{input data}} of ML models consists of five different types (c). The six \textbf{\textit{data-centric tasks}} (d) are applied to the three types of operational data (b) from different stages of the ML pipeline (a) to help people \boxinterpret{\textit{understand}}, \boxdiagnose{\textit{diagnose}}, and \boxrefine{\textit{refine}} ML models.}
 \label{fig:taxonomy}
\end{figure*}

\subsection{Categorization Rationales and Iterations} 
\label{sec:rational}
Our data-centric review was conducted from two aspects: (1) \textit{what} types of data the VIS4ML works focus on; and (2) \textit{how} those data have been operated to interpret, diagnose, or refine ML models. The categorizations of these two aspects have undergone many iterations. We briefly summarize some key iterations here to explain our survey rationales.

For the \textit{``what''} part, we first identified the operational data 
of ML models as \textit{input}, \textit{intermediate}, and \textit{output} data~\cite{zheng2018machine} following the ML execution pipeline (Fig.~\ref{fig:taxonomy}(a, b)).  Then, we tried to label VIS4ML papers based on their interpretation focus across the three data types. However, with some initial labeling, we found that almost all VIS4ML papers covered the \textit{input} and \textit{output} data, some of them used the \textit{intermediate} data while others did not. This categorization quickly degenerated into two categories that essentially reflect if a work is model-specific (using \textit{intermediate} data) or model-agnostic (not using \textit{intermediate} data). As this taxonomy has been introduced in earlier surveys, we did not continue this attempt. Later, we tried to borrow the data categorization from the database field and classified data into \textit{structured} and \textit{unstructured}. With some labeling practices, however, we noticed that most data in VIS4ML works are \textit{unstructured} (e.g., images, texts, and graphs). Using this categorization could not disclose the unique features (e.g., spatial or sequential) of each data type and resulted in a very unbalanced data type distribution. After more explorations and inspired by the underlying data features that ML models are tailored to handle (e.g., CNNs/RNNs are good at processing spatial/sequential data), we eventually came up with our current data categorization (detailed later in Sec.~\ref{sec:data}).

For the ``how'' part, our initial categorization was to group papers based on the VIS techniques they have adopted (e.g., node-link diagrams and scatterplots). This seemed to be the most straightforward choice. However, we soon realized that the identified VIS techniques would be general to any data analysis topics and could not reflect the uniqueness of VIS4ML, nor did they align with our data-centric perspective.
Inspired by Munzner's nested model~\cite{munzner2009nested}, we then shifted our focus to the requirement analysis section of VIS4ML papers. Here, we found that the requirements were mostly task-oriented. 
Therefore, we turned to examine existing VIS task taxonomies, as summarized in Sec.~\ref{sec:relatedsurvey}. Nevertheless, most of those task taxonomies are not specific to VIS4ML but rather general to any data analysis applications.
After several more categorization iterations, we realized that the sentences describing the requirements in individual VIS4ML papers revealed how VIS should serve ML. From those sentences, we extracted the verbs, i.e., operations applied to ML data, and merged similar operations to identify the most representative ones. In the end, we derived six tasks that are specific to VIS4ML (detailed later in Sec.~\ref{sec:task}). Moreover, these tasks are also data-centric, as the objects of the requirement analysis sentences always pertain to the three types of ML operational data. 
To explicitly establish the connections between the identified data and tasks, we connect them with green, orange, and blue arrows between Fig~\ref{fig:taxonomy}(b) and Fig~\ref{fig:taxonomy}(d).

\subsection{Survey Taxonomy and Overview}
\label{sec:taxonomy}
Our data-centric taxonomy reviews VIS4ML papers based on \textit{what} types of data the corresponding ML models focus on and \textit{how} the data have been operated (i.e., VIS4ML tasks) to understand, diagnose, and refine ML models, i.e.,
\begin{itemize}
    \item \textbf{Data Types} (Sec. \ref{sec:data}). We identify the common types of data fed into ML models, describe their unique characteristics, and explain how ML models have been tailored to better learn from them. These data types include: \texttt{tabular}, \texttt{sequential}, \texttt{multi-dimensional array}, \texttt{graph}, and \texttt{multi-modality} data (Fig.~\ref{fig:taxonomy}(c)).
    
    \item \textbf{Data-Centric Tasks} (Sec. \ref{sec:task}). Focusing on the operations applied to the five data types, we elicit six data-centric VIS4ML tasks: \texttt{present}, \texttt{explore}, \texttt{assess}, \texttt{compare}, \texttt{generate}, and \texttt{improve} data. The first five are commonly used for model \boxinterpret{understanding}/\boxdiagnose{diagnosis}.
    The \texttt{generate} task, together with the \texttt{improve} task, is also used for model \boxrefine{refinement} (Fig.~\ref{fig:taxonomy}(d)). 
\end{itemize}

\noindent
\textbf{Overview.} 
Sec.~\ref{sec:data}/Sec.~\ref{sec:task} illustrate our data/task taxonomy in detail, with each sub-category being exemplified by one or multiple representative VIS works. As it is impossible to exemplify all the \fourthnumber{} papers, we summarize them in Tabs.~\ref{tbl:sum} and~\ref{tbl:sum2}. Sec.~\ref{sec:trend} presents the distributions of the papers across data types, data-centric tasks, and their intersections, disclosing the current research trend and prospective future directions. Finally, we discuss some inherent limitations of our survey in Sec.~\ref{sec:discussion} before concluding it in Sec.~\ref{sec:conclusion}.

\section{Data Types}
\label{sec:data}

This section categorizes ML operational data from the input side, as the input data preserve the original characteristics and modality of the data. While we have also considered data categorization using the intermediate or output data, it is important to note that the format of intermediate data is predominantly influenced by the specific ML models employed. For example, in DNNs, the intermediate data are activations and weights, whereas in tree-based models, they become feature-splitting criteria and decision rules. On the other hand, the format of the output data is primarily determined by the addressed applications. For instance, classification and clustering models consistently produce class labels and cluster IDs as outputs, regardless of the input data modalities.
Categorizing VIS4ML papers using the focused ML models/applications has also been covered in early surveys~\cite{jiang2019recent, hohman2018visual}. Our data-centric survey tries to minimize the overlap with them and categorizes data from the input side. Moreover, depending on the specific models and applications, the intermediate and output data can be more complex and diverse compared to the input data. Categorizing papers based on them will require defining and distinguishing numerous subcategories, leading to increased complexity and potential ambiguity in the categorization.

Note that the input data here are the \textit{direct} input to ML models but they may not be the raw data generated from different applications. For example, Tam et al.~\cite{tam2011visualization} studied facial dynamics data to analyze the difference between four facial emotions, anger, surprise, sadness, and smile. The raw data are videos captured from the face of different people, but these videos cannot be directly used to train ML models. The authors pre-processed individual video frames first to extract 14 numerical measurements for different facial features, e.g., the vertical displacement of the chin. Tracking the values of these measurements across frames forms a time-series that can be fed into ML models. In this case, the input data are the time-series rather than the raw videos.

Based on a comprehensive review of the \fourthnumber{} papers, we categorize the input of ML models into the following five types: \texttt{tabular}, \texttt{sequential}, \texttt{multi-dimensional array}, \texttt{graph}, and \texttt{multi-modality} data. All data types come with a collection of instances, and each instance may have some annotation information associated with it. 
Mathematically, a dataset $\mathcal{D}$ can be described as:
\begin{equation}
\begin{array}{c}
\mathcal{D} = {<}X, Y{>},\ where \\
X {=} \{x_1, x_2, ..., x_n\}\  and \ Y {=} \{y_1, y_2, ..., y_n\}.
\end{array}
\label{eq:data}
\end{equation}
$X$ is the feature part of $\mathcal{D}$, which is the input of ML models. 
The term ``feature'' has the same meaning as in ML, i.e., it denotes ``an individual measurable property''~\cite{chandrashekar2014survey}, e.g., the \textit{age}, \textit{gender}, or \textit{annual income} of an individual. $Y$ (if exists) is the annotation part that supervised ML models should target on during training, e.g., the class labels of image data. 
As $X$ and $Y$ have the one-to-one correspondence, a single instance of $\mathcal{D}$ can be denoted as $(x_i, y_i)$. In cases where $\mathcal{D}$ does not contain $Y$, ML models will have to learn from $X$ in an unsupervised or semi-supervised manner. 

The differences of the five data types reside in the $X$ part. 
We explain them in the following subsections by (1) providing their definition, (2) listing some typical examples, and (3) discussing the challenges when learning from them.
\subsection{Tabular Data}
\setlength\intextsep{0pt}
\begin{wrapfigure}[4]{L}{0.15\columnwidth}
\includegraphics[width=42pt]{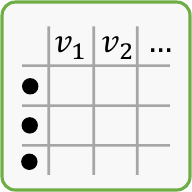}
\end{wrapfigure}
\textbf{\textit{Definition.}} \texttt{Tabular} data comes as a data table, where each row is a data instance and each column is an attribute of the instance. Mathematically, a/an row/instance can be denoted as:
\begin{equation}
\label{eq:table}
    x_i=(v^{D_1}, v^{D_2}, ..., v^{D_m}),
\end{equation}
where $v^{D_j}$ ($j{\in}[1,m]$) is a possible value of the $j$th attribute defined in the corresponding domain $D_j$, and the value can be either categorical or numerical. The annotation information $Y$, if exists, usually appears as a column in the table.

\textbf{\textit{Examples.}} The U.S. Census Income dataset used in~\cite{wexler2019if} is a typical tabular data. Each row of the dataset is a person and each column reflects the value of one feature, e.g., \textit{age}, \textit{gender}, and \textit{capital-gain}. Similar examples also include the Bank Marketing dataset used in~\cite{ming2018rulematrix}, and the Criminal Recidivism dataset used in~\cite{kaul2021improving}. Individual features of these tabular data usually represent human-understandable semantics, e.g., \textit{age}, \textit{race}, and \textit{income}, which contribute significantly to the interpretation of the corresponding ML models. Moreover, new features can also be generated through feature engineering to horizontally extend the table.

\textbf{\textit{Challenges.}} The key challenge for ML models in handling tabular data is to manage the large number of features and learn information out of their complicated collaborative effects, i.e., feature interactions~\cite{molnar2020interpretable}.
Both traditional ML models (e.g., SVMs, logistic regressions, decision trees) and DL models (e.g., multi-layer perceptions) have been applied to this type of data. VIS4ML works have covered all these models' interpretations~\cite{wexler2019if,kaul2021improving, hohman2019telegam, cabrera2019fairvis, cheng2020dece} with varying visualization focuses, such as interpreting these models by better presenting individual instances~\cite{krause2017workflow}, more intuitively disclosing the importance of features~\cite{cheng2020dece}, and steering the feature engineering process to refine these models~\cite{chatzimparmpas2022featureenvi}.

\subsection{Sequential Data}
\setlength\intextsep{0pt}
\begin{wrapfigure}[4]{L}{0.15\columnwidth}
\includegraphics[width=42pt]{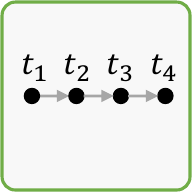}
\end{wrapfigure}
\textbf{\textit{Definition.}} \texttt{Sequential} data comes with a collection of sequences that may have varying lengths. Each sequence $x_i$ is composed of $k$ tokens organized in order. For example, a sentence with $k$ words is a sequence of $k$ tokens. Each token $t_i$ is a feature vector, e.g., the embedding vector of a word. Mathematically,
\begin{equation}
\label{eq:sequence}
x_i = (t_1, t_2, ..., t_k).  
\vspace{-0.021in}
\end{equation}
Note that we used $x_i$ in both Eq.~\ref{eq:table}, Eq.~\ref{eq:sequence}, and later equations, to denote a single instance of the dataset $X$. However, it has different representations when the data type is different.

\textbf{\textit{Examples.}} The two most common sequential data are text data (each word/character is a token) and time-series data (each time step is a token). For example, the Penn TreeBank~\cite{marcus1993building} dataset used in~\cite{ming2017understanding, strobelt2017lstmvis} is a famous English corpus of sentences. Each sentence is a sequential instance and the parts of speech for individual words/tokens have been well-annotated in the dataset.
Weather forecasting data~\cite{roesch2019visualization}, sleep signals~\cite{garcia2019v}, and musical chord progression sequences~\cite{strobelt2017lstmvis} are examples of time-series data, in which, tokens are ordered into sequences chronologically.

\textbf{\textit{Challenges.}} The main challenge of learning from sequential data is to capture the sequential information propagation inside a sequence and find how preceding and succeeding tokens influence each other. RNNs and their variants (e.g., LSTMs and GRUs~\cite{ming2017understanding}) that maintain multiple hidden states to recursively pass on the sequential information from token to token demonstrate superior performance on this data type. Recently, Transformers~\cite{vaswani2017attention} have also been introduced for sequential data learning. Instead of processing the tokens sequentially one-by-one, Transformers consume all tokens at once and use the self-attention mechanism to learn pair-wise attentions between all tokens. 
Most VIS4ML works for this data type focus on presenting the sequential data and relating them with their latent representations inside ML models to reveal what the models have captured, e.g., RNN hidden state interpretations~\cite{strobelt2017lstmvis,karpathy2015visualizing}. Explaining how Transformers' self-attentions work so well on sequential data has also been extensively conducted~\cite{park2019sanvis,derose2020attention}.

\subsection{Multi-Dimensional (MD) Array Data}
\setlength\intextsep{0pt}
\begin{wrapfigure}[4]{L}{0.15\columnwidth}
\includegraphics[width=42pt]{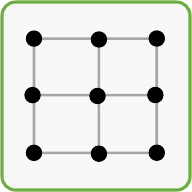}
\end{wrapfigure}
\textbf{\textit{Definition.}} \texttt{Multi-dimensional array} data is composed of a set of instances, each is an array of scalar values organized spatially into a regular grid structure. For example, a gray-scale image can be considered as a 2D array storing the image's pixels along the width and height dimensions. Using \textit{multi-dimensional (MD) array} to name this type of data follows the terminology from the ML domain, i.e., LeCun et al.~\cite{lecun2015deep} and Goodfellow et al.~\cite{goodfellow2016deep} referred to this type of data as ``multiple arrays'' and ``multidimensional arrays,'' respectively. Mathematically, each instance can be denoted as (assuming a 2D case),

\begin{equation*}
x_i = 
\begin{pmatrix}
s_{1,1} & s_{2,1} & \cdots & s_{w,1} \\
s_{1,2} & s_{2,2} & \cdots & s_{w,2} \\
\vdots  & \vdots  & \ddots & \vdots  \\
s_{1,h} & s_{2,h} & \cdots & s_{w,h} 
\end{pmatrix}.
\end{equation*}

\textbf{\textit{Examples.}} Image and volume data are representative examples for this data type. For instance, the MNIST dataset~\cite{deng2012mnist} used in~\cite{wang2018ganviz, rauber2016visualizing} is a famous benchmark, consisting of 70,000 gray-scale images of hand-written digits. Each image/instance is a 2D array with individual scalar values (pixels) ranging from 0 to 255. The CIFAR10~\cite{krizhevsky2009learning} used in~\cite{liu2016towards, liu2017analyzing} and the ImageNet~\cite{russakovsky2015imagenet} used in~\cite{hohman2019s, das2020bluff} are RGB image datasets with higher-resolution images in more classes (each image is a 3D array of scalar values).

\textbf{\textit{Challenges.}} Preserving spatial continuity and extracting localized features are the essential challenges for ML models when learning from MD-array data. CNNs~\cite{lecun1998gradient} are often the ideal choices in handling MD-array data, as they can chain layers of convolutional filters to extract varying features hierarchically (e.g., the basic shape/color features from lower CNN layers and the complicated objects/concepts from higher layers). Lately, vision Transformers~\cite{dosovitskiy2020image} and their combinations with CNNs have also demonstrated outstanding performance on this type of data.
VIS4ML works strive to better demonstrate the spatial features of MD-array data~\cite{bertucci2022dendromap}, highlight important features impacting ML models' behaviors, (e.g., salience map visualizations~\cite{wang2021visual}), and externalize the internal representation of the data inside ML models (e.g., feature map visualizations~\cite{wang2020cnn}).

\subsection{Graph Data}
\setlength\intextsep{0pt}
\begin{wrapfigure}[4]{L}{0.15\columnwidth}
\includegraphics[width=42pt]{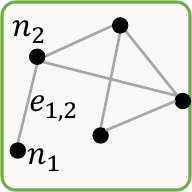}
\end{wrapfigure}
\textbf{\textit{Definition.}} A \texttt{graph} is usually represented by a set of nodes and a set of edges. 
The nodes contain feature information and the edges record the connections between nodes. Formally, a graph can be denoted as,
\begin{equation}
\begin{array}{c}
    \mathcal{G} = {<}\mathcal{N}, \mathcal{E}{>},\\ 
    where\ \mathcal{N} {=} \{n_1, n_2, ..., n_n\}, \mathcal{E} {=} \{e_{i, j}| i{<=}n, j{<=}n\}.
  \end{array}
\end{equation}
Each node is further represented by a feature vector, i.e.,
\begin{equation}
\label{eq:graphnode}
    n_i=(f_1, f_2, ..., f_k).
\end{equation}
In general, graph data are often categorized into \textit{homogeneous} and \textit{heterogeneous} graphs. For the former, all graph nodes represent instances of the same type and all graph edges denote the same relationship between nodes. For the latter, however, the graph nodes have varying types and the graph edges could represent multiple relationships. 

\textbf{\textit{Examples.}} A social network is a typical homogeneous graph, where each node is a person and each edge reflects the friendship between persons. Each person will also have multiple features, e.g., \textit{gender}, \textit{age}, \textit{number of friends}, etc., constituting the feature vector of the corresponding graph node. More homogeneous graph examples include publication citation graphs~\cite{mccallum2000automating} and molecular compound structure graphs~\cite{morris2020tudataset}. For heterogeneous graphs, the User-Movie data used in~\cite{liu2022visualizing} is a good example, where a graph node could either be a user or a movie, and an edge between two nodes represents the user has watched the corresponding movie.

\textbf{\textit{Challenges.}} ML models can be trained to learn from both the node-related features and the edge-related structures of graphs. Often, the training instances fed into ML models are individual nodes, each is represented by a feature vector (Eq.~\ref{eq:graphnode}). The ML models learn from these nodes' features, as well as the features from their neighboring nodes through edge connections, to predict the properties of certain nodes or the existence of specific edges.
Accordingly, the challenge in handling graph data is to not only learn from the features of individual nodes, but also leverage their neighbors' features that can be propagated to them through connected edges (i.e., learning from both the feature and structure information). GNNs~\cite{wu2020comprehensive} are introduced to take care of the message passing between nodes, as well as the aggregation of information received from a node's neighbors. Their power has been demonstrated across all types of graph-related learning tasks, e.g., node classification, node ranking, edge prediction, and community detection. 
The difficulties that VIS4ML faces with this type of data are to effectively present the multivariate features of graph nodes (e.g., glyph visualization~\cite{jin2022gnnlens}), disclose the sophisticated connections between nodes, and more importantly, address the scalability issues when the graphs become large.

Note that there are also ML models designed to learn from multiple graphs. In this case, each training instance is a graph (rather than a graph node). Individual graphs have their independent sets of nodes and edges. For example, a chemical compound can be represented as a graph (node: atom, edge: bond). Researchers have developed many DL models (i.e., binary classifiers) to predict if a compound is cancer-related or not~\cite{wale2008comparison}. Uniformly handling the varying graph sizes and efficiently extracting information out of individual graphs are the key learning challenges.

\subsection{Multi-Modality Data}
\setlength\intextsep{0pt}
\begin{wrapfigure}[4]{L}{0.15\columnwidth}
\includegraphics[width=42pt]{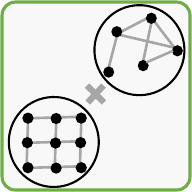}
\end{wrapfigure}
\textbf{\textit{Definition.}} Multiple of the aforementioned data types could be learned together by ML models. These data may come from different data sources, be in different formats, and present different modalities. We call them \texttt{multi-modality} data, and their modalities could be nested or interwoven.

\textbf{\textit{Examples.}} Video data can be considered as a hybrid of MD-array and sequential data. Each frame of the video is an image encoding spatial features. A consecutive sequence of these frames constitute a sequential data instance. The spatial modality is nested inside the sequential modality. 
Most of the deep reinforcement learning (DRL) agents trained to play video games use this type of multi-modality data as training instances (i.e., game episodes)~\cite{wang2021visual,jaunet2020drlviz}. Dynamic graphs hybrid sequential data with graph data, and the graph modality is nested under the sequential modality, e.g., an evolving social network with varying numbers of nodes (users) and edges (users' relationships) over time.
Different modalities can also be interwoven at the same level. For example, the data used in $\textit{M}^2$Lens~\cite{wang2021m2lens} include three types of sequential data with different modalities, (1) facial expressions (video data), (2) voices of speakers (acoustic data), and (3) verbal transcripts (text data). Different ML models can be trained to take care of the respective modalities and their outcomes can be fused together for comprehensive learning.

\textbf{\textit{Challenges.}} The challenges of learning from this type of data come from choosing the best ML models to handle individual data modalities and effectively fusing the learned outcomes.
Different ML models are good at handling different data types. For example, tree-based models take good care of the feature interactions of tabular data; CNNs are good at extracting spatial features from MD-array data; RNNs show superior performance in managing data with sequential structures; GNNs demonstrate advantages in capturing the structure-level information of graphs. How to integrate these ML models and maximally leverage their respective advantages to process the multi-modality data is a challenging problem and of paramount importance.
VIS4ML strives to better visualize individual modalities of the data and effectively reveal the underlying connections between modalities. Furthermore, the complicated relationship between varying modalities also challenges VIS4ML works to take advantage of the hidden information between modalities to refine and improve ML models~\cite{jaunet2021visqa,chen2021towards}.

\section{Data-Centric VIS4ML Tasks}
\label{sec:task}
As summarized in earlier works~\cite{choo2018visual, liu2017towards}, VIS has served ML in model understanding, diagnosis, and refinement. 
To analyze how these goals are achieved from the data side, we investigate the concrete VIS tasks that have been conducted on the \textit{input}, \textit{intermediate}, and \textit{output} data (Fig.~\ref{fig:taxonomy}(b)). The six elicited tasks are: \texttt{present}, \texttt{explore}, \texttt{compare}, \texttt{assess}, \texttt{generate}, and \texttt{improve} data. Their relationship with model \boxinterpret{understanding}, \boxdiagnose{diagnosis}, and \boxrefine{refinement} is reflected in Fig.~\ref{fig:taxonomy}(d). Note that some of the tasks have been covered in earlier surveys, e.g., \texttt{present} and \texttt{compare}. Here, we focus on illustrating how they have been applied to the operational data in the VIS4ML context. There are also tasks that are not well-covered in earlier surveys, e.g., \texttt{generate} and \texttt{improve}. These are specific tasks identified from our data-centric review of the literature.

\subsection{Present Data}
Presenting data is to map the operational data into different visual channels to externalize the information in the data. It is a fundamental VIS operation that every VIS4ML work conducts, but different works may focus on the data from different ML pipeline stages. As the data to ML models is a collection of instances (Eq.~\ref{eq:data}), the visual mappings focus either on individual data instances or on the aggregation of a group of instances (instance/group-level). We thus explain the \texttt{present} task from these two levels. Inside each, we use some typical VIS4ML works to explain how individual \textit{input}, \textit{intermediate}, and \textit{output} data instances/groups have been presented. For a full list, please refer to Tabs.~\ref{tbl:sum} and~\ref{tbl:sum2}.

\subsubsection{Instance-Level Data Presentation}
Instance-level presentation visually encodes the information of individual data instances. Users can directly interact with each instance (if needed) to examine ML models' behaviors. 

\begin{figure}[tbh]
 \centering 
 \includegraphics[width=\columnwidth]{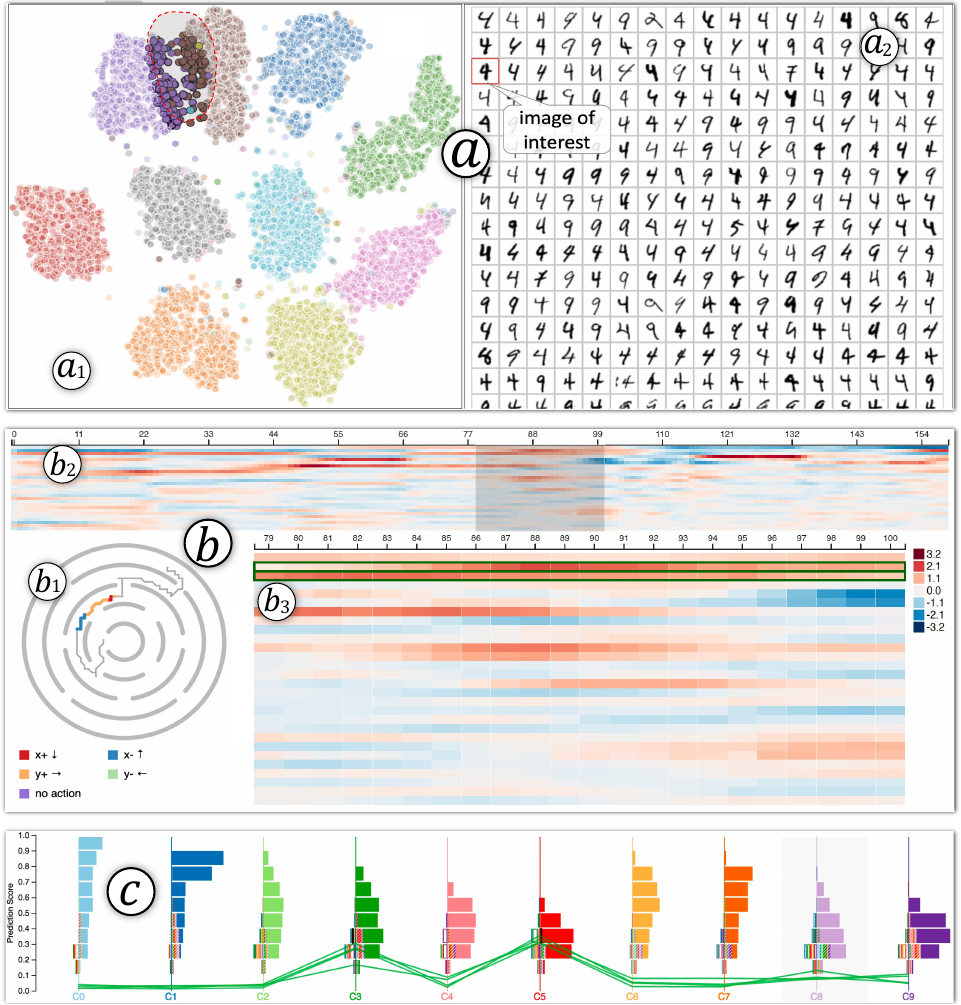}
  \vspace{-0.3in}
 \caption{Instance-level presentation. (a) The \textit{input} images are presented through a scatterplot, one point for one image. Image courtesy of Wang et al.~\cite{wang2019deepvid} \copyright \ 2019 IEEE. (b) The \textit{intermediate} hidden states are externalized through a heatmap, each row is an instance (principle component) and each column is a time step. Image courtesy of He et al.~\cite{he2020dynamicsexplorer} \copyright \ 2020 IEEE. (c) The \textit{output} probabilities are presented through a PCP, one polyline per instance. Image courtesy of Ren et al.~\cite{ren2016squares} \copyright \ 2016 IEEE. }
 \vspace{-0.2in}
 \label{fig:instance}
\end{figure}

\textit{Presenting Input Data.} 
Individual input instances carry data features/semantics that are important to understand the behavior of ML models. Presenting input instances of interest is therefore the starting point of many VIS4ML works. For example, DeepVID~\cite{wang2019deepvid} presents the MD-array input of a classification model as a grid of images (Fig.~\ref{fig:instance}($\text{a}_2$)). From the visual appearance of the images, users can select the ones that are more likely to confuse the classifier to diagnose the model. As directly visualizing all input images in the grid will have a severe scalability issue, the authors use the images' extracted features to present an overview of them first before the grid layout. Specifically, a pre-trained CNN is used as a feature-extractor to extract the essential features of the input images. These HD features are then reduced to 2D through dimensionality reduction~\cite{van2008visualizing} and visualized as a scatterplot (Fig.~\ref{fig:instance}($\text{a}_1$)). Each point in the plot represents one input image and it is colored by its class label. 
From such an overview, images that are similar to both digit 4 and 9 can be easily selected to probe the classifier's decision boundary between these two classes. 
Note that the extracted features of the input images (from the third-party CNN) are not the intermediate data of the interpreted classifier and DeepVID is a model-agnostic interpretation method.

\textit{Presenting Intermediate Data}. Intermediate data is the key to opening ML black-boxes~\cite{cashman2018rnnbow,derose2020attention}. Heatmap is commonly used for its visualization, which presents data through a 2D matrix and uses the color of each matrix cell to encode the information. For example, DynamicsExplorer~\cite{he2020dynamicsexplorer} adopts a heatmap (Fig.~\ref{fig:instance}(b)) to investigate an LSTM-based DRL agent trained for the ``ball-in-maze'' game (Fig.~\ref{fig:instance}($\text{b}_1$)). To better handle the high-dimensionality of the intermediate hidden states, PCA is applied onto the hidden states first. In Fig.~\ref{fig:instance}($\text{b}_2$), the horizontal and vertical axes of the heatmap represent time and individual principle components, respectively. Users can brush horizontally to select the interested temporal range and examine the hidden states (Fig.~\ref{fig:instance}($\text{b}_3$)).

\begin{figure*}[tbh]
 \centering 
 \includegraphics[width=\textwidth]{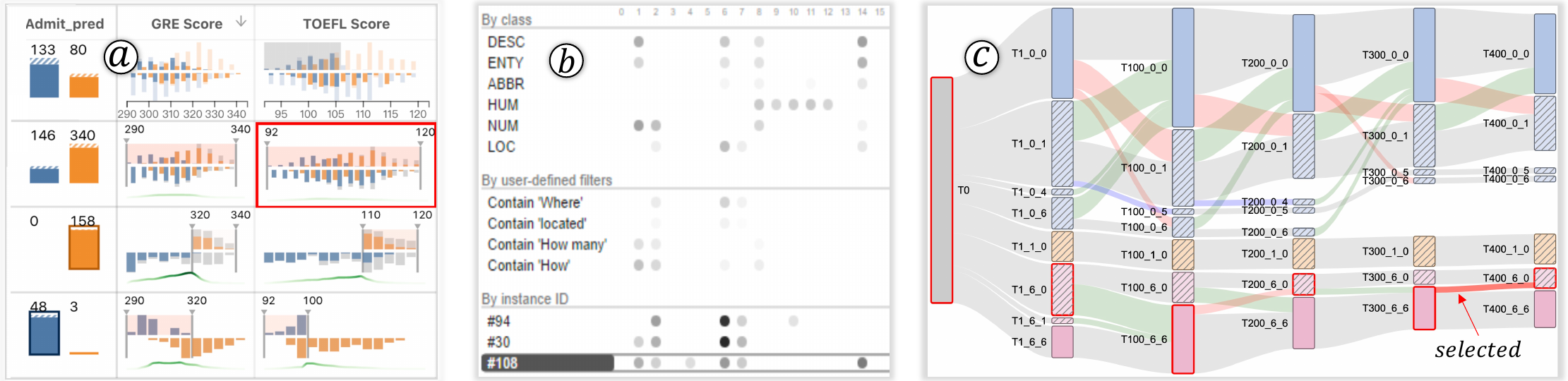}
  \vspace{-0.3in}
 \caption{Group-level presentation. (a) The tabular \textit{input} data in DECE~\cite{cheng2020dece} are divided into subgroups and presented as histograms. Image courtesy of Cheng et al.~\cite{cheng2020dece} \copyright \ 2020 IEEE. (b) The \textit{intermediate} DNN activations from subgroups of instances are aggregated in ActiVis~\cite{kahng2017cti} and presented as circles, whose color denotes the active level. Image courtesy of Kahng et al.~\cite{kahng2017cti} \copyright \ 2017 IEEE. (c) A Sankey-diagram based temporal confusion matrix is used to present the \textit{output} prediction over the training of a tree-boosting model.
 Image courtesy of Wang et al.~\cite{wang2021investigating} \copyright \ 2021 IEEE.}
 \vspace{-0.1in}
 \label{fig:group}
\end{figure*}

\textit{Presenting Output Data}. Parallel coordinates plots (PCPs) have been used widely to present the output of ML models. For example, Ren et al.~\cite{ren2016squares} employ a PCP to visualize the prediction probabilities from a classification model. As shown in Fig.~\ref{fig:instance}($\text{c}$), each parallel axis denotes one class and the values on it show the predicted probabilities for the corresponding class. A polyline connecting the probabilities across classes shows the entire output probability distribution for an instance. Multiple instances are presented as multiple superimposed polylines, and their collective behaviors reveal the model's performance over classes. In Fig.~\ref{fig:instance}($\text{c}$), four MNIST images with similar probabilities to be digit '3' and '5' are shown as four polylines in the PCP.

\subsubsection{Group-Level Data Presentation}
Group-level presentation first aggregates data instances into groups and then visually encodes them. It focuses more on revealing group-level data patterns, instead of disclosing individual instances' local behaviors.

\textit{Presenting Input Data.} Histogram is a popular VIS technique to present data distribution across input feature values. For example, DECE~\cite{cheng2020dece} uses a big table of histograms to present the tabular data fed into ML models. As shown in Fig.~\ref{fig:group}(a), each row of the table is a subgroup of instances and each column is a data feature. The upward histogram in each table cell presents the distribution of the corresponding feature values (for the subgroup of instances). 
Based on the binary prediction results of the instances, the upward histogram is further divided into two juxtaposed ones, colored by blue and orange.
Moreover, the counterfactual examples for the subgroup of instances are also generated and their feature value distributions are presented as a symmetric but downward histogram. The side-by-side comparison helps users formulate/verify hypotheses on different features.

\textit{Presenting Intermediate Data.} Matrix visualization aggregates data instances into a 2D matrix and uses colors, sizes, or glyphs to encode the aggregated data inside each cell. For example, ActiVis~\cite{kahng2017cti} enables users to flexibly define data subgroups, e.g., by class labels. 
Aggregating the instances' activations from a DNN inside individual subgroups and comparing the aggregated activations across subgroups disclose the functionality of different DNN neurons. As shown in Fig.~\ref{fig:group}(b), each row/column of the matrix represents a subgroup/a neuron, and the circle inside a cell represents the aggregated response-level of the corresponding neuron (darker colors indicate stronger aggregated responses).

\textit{Presenting Output Data.} Sankey-diagram can effectively illustrate how data instances are divided or merged into groups (often over time) and is a common technique for group-level data presentation. For example, VISTB~\cite{wang2021investigating} employs a Sankey-diagram to disclose the evolution of predictions over the training of a tree-boosting model. As shown in Fig.~\ref{fig:group}(c), each column of nodes presents the confusion matrix of the model at a time step. The color and filling pattern denote the predicted class and prediction correctness (solid: true positive (TP); strip: false positive (FP)), respectively. The bands between neighboring columns illustrate the flowing of instance groups between time steps. 
Their color reflects if the predictions of the corresponding groups are improved (green: from a FP to a TP cell), degenerated (red: from a TP to a FP cell), or not changed (gray). Such a visualization effectively monitors the model's performance evolution.

\subsection{Explore Data}
Visual data exploration is ``an undirected search for relevant information within the data''~\cite{tominski2006event}, in which users may not have a clear goal while playing with the data but rely on highly interactive interfaces and intermediate insights to drive the exploration. In VIS4ML, when data gets too large and/or contains multiple facets, explorations will have to come into the picture. Based on the exploration directions, we organize works into \textit{vertical} and \textit{horizontal} explorations. 

\subsubsection{Vertical Exploration}
Vertical exploration refers to the process of exploring data by following the order of either global-to-local (top-down) or local-to-global (bottom-up). The former starts by providing users with a succinct data overview, from which, users can drill down to low-level data details on-demand. In contrast, the latter first investigates part of the data locally with sufficient details. Based on the knowledge obtained from some representative data instances/features, the users then expand the exploration to the entire dataset.

The {\textit{top-down}} exploration follows Shneiderman's information seeking mantra~\cite{shneiderman2003eyes} to present data through \textit{overview +details}. 
For example, DeepVID~\cite{wang2019deepvid} diagnoses incorrect predictions of image classifiers by first laying out all images using tSNE+scatterplot. The layout provides an overview of all images, guiding users to drill down to individual images of interest for detailed diagnosis. 
As shown in Fig.~\ref{fig:instance}($\text{a}_1$), the user selects the instances between the purple and brown clusters (e.g., images with similar probabilities to be digit '4' and '9') through a lasso selection. Fig.~\ref{fig:instance}($\text{a}_2$) presents the details of these images and enables the user to further investigate individual ones.
Similarly, VATLD~\cite{gou2020vatld} lays out all images through a performance landscape, i.e., \textit{TileScape}, for an overview. Each tile aggregates similar images and uses the instance with the median score to represent the tile. Interactive zooming empowers users to explore the space and drill down to finer data granularities on-demand.

The {\textit{bottom-up}} exploration inspects individual instances first, and then, expands the inspections to all instances to augment the findings. For example, LSTMVis~\cite{strobelt2017lstmvis} allows users to interactively define the active pattern of different LSTM hidden states through an on-off curve defined over a single instance. 
The pattern is then used as a template to match with all instances. From the semantics augmented by all matched instances, the authors confidently interpret what has been captured by different hidden states.
DQNViz~\cite{wang2018dqnviz} closely examines how a DRL agent plays an Atari game in one game episode and uses a regular expression to define its playing strategy. The regular expression is then applied to all game episodes to search when and where the same strategy was used to understand the agent's behaviors.

\begin{figure}[tb]
 \centering 
 \includegraphics[width=\columnwidth]{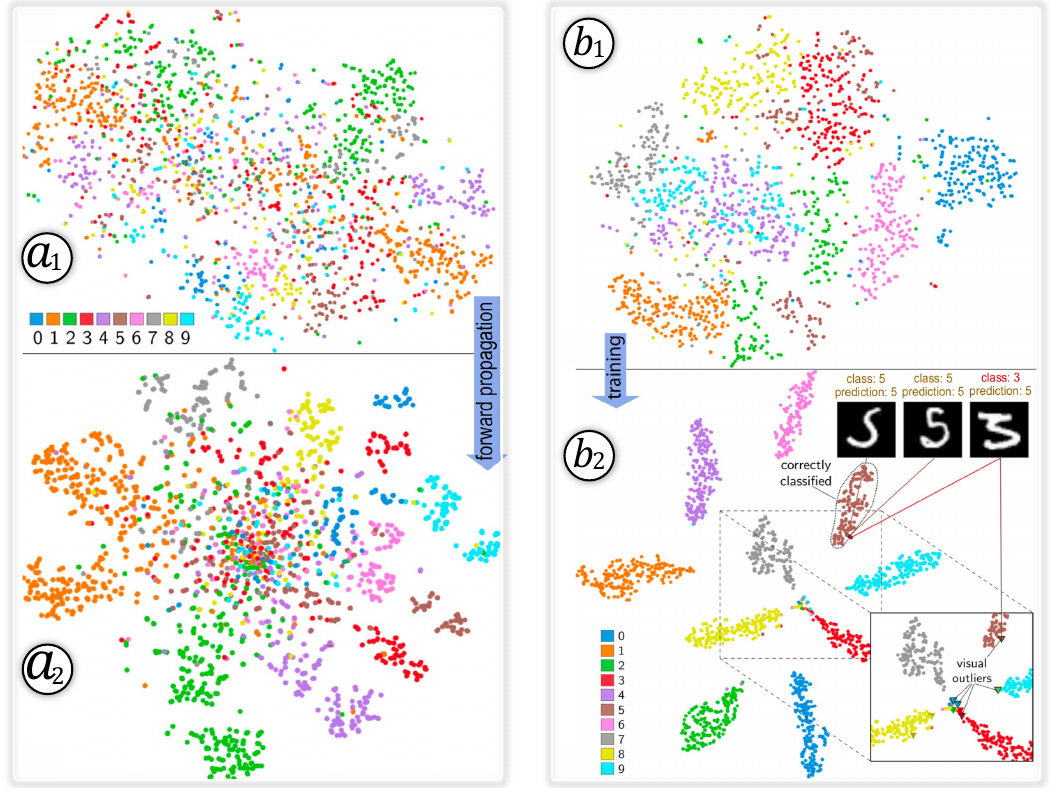}
  \vspace{-0.3in}
 \caption{Explore data across different DNN layers (a1, a2) or training iterations (b1, b2). Image courtesy of Rauber et al.~\cite{rauber2016visualizing} \copyright \ 2016 IEEE.}
 \label{fig:horizontal}
\end{figure}
\subsubsection{Horizontal Exploration}
\begin{figure*}[tbh]
 \centering 
 \includegraphics[width=\textwidth]{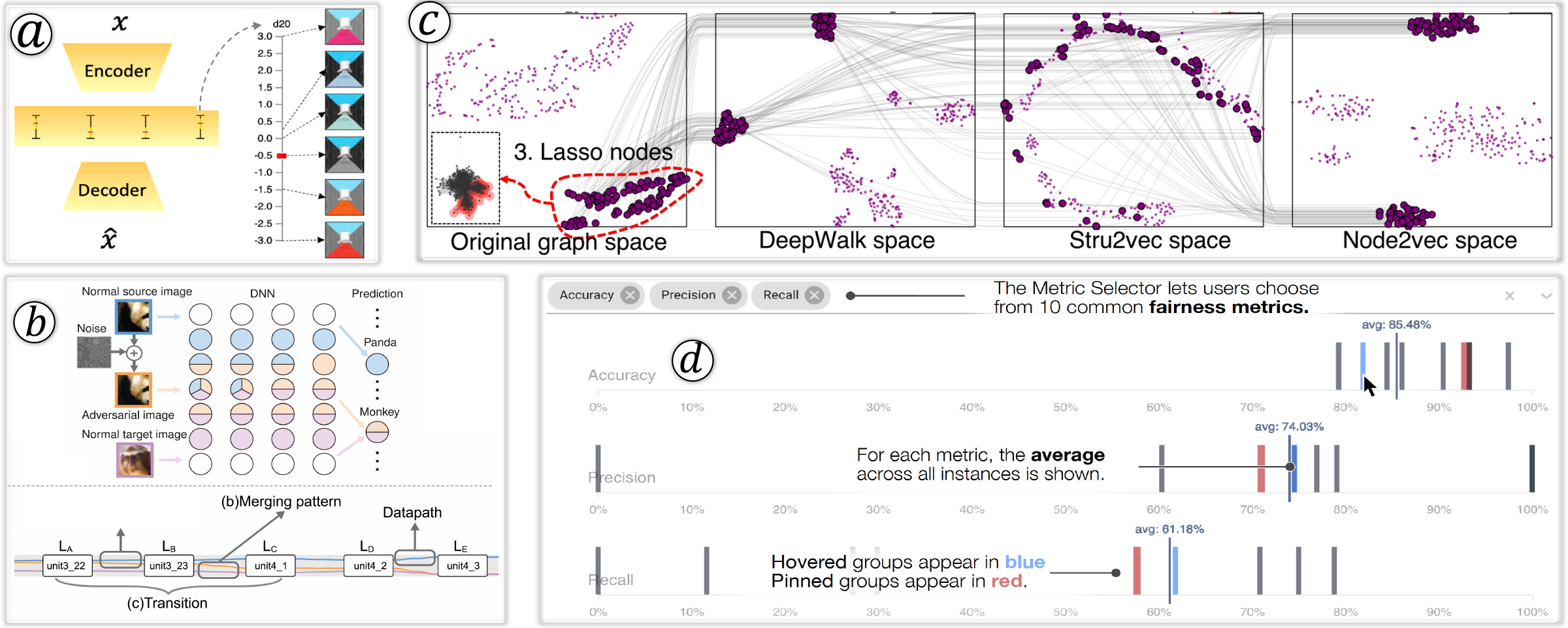}
  \vspace{-0.3in}
 \caption{Comparison:
 (a) Intra-instance: SCANViz compares the reconstructions of the same image. Image courtesy of Wang et al.~\cite{wang2020scanviz} \copyright \ 2020 IEEE. (b) Inter-instance: AEVis compares the datapaths of three images to diagnose adversarial attacks. Image courtesy of Cao et al.~\cite{cao2020analyzing} \copyright \ 2020 IEEE. (c) Intra-group: EmbeddingVis compares the embeddings for the same group of instances from different models. Image courtesy of Li et al.~\cite{li2018embeddingvis} \copyright \ 2018 IEEE. (d) Inter-group: FairVis compares model performance across instance groups. Image courtesy of Cabrera et al.~\cite{cabrera2019fairvis} \copyright \ 2019 IEEE.}
 \vspace{-0.1in}
 \label{fig:comparison}
\end{figure*}

Horizontal exploration explores data across multiple stages of the ML pipeline, multiple temporal iterations, or multiple data spaces to relate data and derive insights. For example, Rauber et al.\cite{rauber2016visualizing} employ tSNE+scatterplot to visualize the activations of all data instances from early and later layers of a DNN, as shown in Fig.~\ref{fig:horizontal}($\text{a}_1$, $\text{a}_2$).
The two layouts clearly disclose how the forward-propagation separates data instances into different classes. Similarly, Fig.~\ref{fig:horizontal}($\text{b}_1$, $\text{b}_2$) show the layouts for the DNN's last-layer activations from two training stages. Exploring these visualizations helps to understand the model's temporal evolution.
DGMTracker~\cite{liu2017analyzing} explores deep generative models layer-by-layer through statistics presented by line-chart snapshots to diagnose the model training process. The exploration traces data across neural network layers sequentially, which is considered a horizontal exploration. 
EmebeddingVis~\cite{li2018embeddingvis} simultaneously explores multiple graph embedding spaces generated for the same set of graph nodes by using different embedding algorithms. As shown in Fig.~\ref{fig:comparison}(c), the original graph space and three embedding spaces are presented as four juxtaposed scatterplots. Explicit links are used to connect the same graph nodes across spaces for coordinated explorations, which facilitates the comparison of the underlying embedding algorithms. Specifically, the DeepWalk and Node2vec algorithms perform similarly well in separating the selected nodes (in the red dashed line) into two subgroups, whereas the Stru2vec algorithm disperses them.

\subsection{Compare Data}
Data comparisons in VIS4ML identify the similarity and difference of the operational data to support model understanding or diagnosis. 
They focus either on individual data \textit{instances} or \textit{groups} of instances, and the comparisons are often conducted either \textit{within} or \textit{between} instance(s)/group(s). 

\subsubsection{Intra-Instance Comparison}
The intra-instance comparison compares the same data instance before and after some modifications applied to either the data instance or the studied ML model.

For the first case (fix model, modify data), researchers modify a single data instance and examine how the modification impacts the ML model to probe its behavior. 
For example, SCANViz~\cite{wang2020scanviz} uses a PCP to present the latent dimensions of a $\beta$VAE trained on images (Fig.~\ref{fig:comparison}(a)). By perturbing the value of a latent dimension and interactively decoding the perturbed latent representations back as images, users can conclude what the dimension has encoded. Specifically, the six images in Fig.~\ref{fig:comparison}(a) show six reconstructions of the same input image, but with different values on dimension 20. By comparing them, we can see this latent dimension majorly controls the \textit{floor color} of the 3D scene. More intra-instance comparisons include the works built upon what-if analyses and counterfactual examples~\cite{wexler2019if, cheng2020dece}, which often perturb the input features of tabular data.

For the second case (fix data, modify model), the data instance is intact but its intermediate/output representations become different due to model modifications. Comparing the instance's intermediate/output representations reveals the corresponding model's evolution. For example, Attention Flows~\cite{derose2020attention} introduces a radial layout to compare the self-attention of a Transformer model on a sentence (a sequential data instance) before and after the model's fine-tuning. The comparison helps to understand how the fine-turning process adapts the model to the data.

\subsubsection{Inter-Instance Comparison}
Inter-instance comparison compares two or more instances, generating model insights based on the model's dissimilar behaviors on them. For example, AEVis~\cite{liu2018analyzing,cao2020analyzing} interprets how an adversarially generated \textit{panda} image was incorrectly predicted as a \textit{monkey} by comparing the datapaths of a normal \textit{panda} image, its adversarial counterpart, and a normal \textit{monkey} image. As shown in Fig.~\ref{fig:comparison}(b), the three colors, blue, orange, and purple, correspond to the neurons that are activated by the three images, respectively. Connecting neurons of the same color across layers forms the datapath for the corresponding image. 
The authors also design a new visualization to effectively present these datapaths and their evolution patterns over time (Fig.~\ref{fig:comparison}(b), bottom). Comparing the datapaths of the three images, especially where the datapath of the adversarial image diverges from the \textit{panda} and merges into the \textit{monkey}, helps to locate where the adversarial attack happens.
Similarly, GANViz~\cite{wang2018ganviz} compares a pair of real and generated images from a GAN model to study how its discriminator works in the adversarial settings.

Note that some interpretation methods may fall into both intra-instance and inter-instance comparison based on how the comparison was conducted. For example, when interpreting ML models with counterfactual examples, the examples could be generated by perturbing a single data instance of interest. Only one instance is involved in this case and the work belongs to our ``intra-instance'' comparison category. Nevertheless, there are also works generating counterfactual examples by searching from the existing data instances. In this case, two or more instances will be involved and it falls into our ``inter-instance'' comparison category.

\subsubsection{Intra-Group Comparison}
The intra-group comparison in VIS4ML either (1) compares different models' performance using the same group of instances for a fair evaluation; or (2) compares the same group of instances at different stages of a model to understand its evolution. For case (1), EmbeddingVis~\cite{li2018embeddingvis} compares different embeddings of the same set of graph nodes generated from different embedding algorithms. As shown in Fig.~\ref{fig:comparison}(c), each scatterplot shows the dimensionality reduction result for the embedding generated by one algorithm. Embeddings from different algorithms are comparable since they are for the same set of instances. Also, there are one-to-one correspondences across the embeddings, as reflected by the curves connecting the instances across plots. 
For case (2), Xiang et al.~\cite{xiang2019interactive} propose DataDebugger to interactively correct input data with incorrect labels over multiple iterations. In each iteration, the distribution of data instances and their prediction statistics are presented through the proposed incremental tSNE. Comparing the distributions and statistics for the same group of instances across iterations discloses the data quality improvement over time.

\subsubsection{Inter-Group Comparison}
Inter-group comparison divides data into subgroups and compares the behavior discrepancy among the subgroups to interpret ML models.
For example, ActiVis~\cite{kahng2017cti} interprets DNNs by allowing users to flexibly define instance groups (e.g., misclassified instances with common features) and aggregate the activations of the same group for cross-group comparisons (explained in Fig.~\ref{fig:group}(b)). FairVis~\cite{cabrera2019fairvis} compares the performance across subgroups of instances with different features to disclose the biases hidden in predictive models. As demonstrated in Fig.~\ref{fig:comparison}(d), each row of strip plot presents the studied model's performance with one metric (e.g., accuracy, precision, and recall), and each strip bar (inside a row) represents one subgroup.
In the top row, the red \textit{Female} group has 10\% more accuracy than the blue \textit{Male} group, indicating potential gender discrimination.
To investigate how adversarial attacks work in CNNs, Bluff~\cite{das2020bluff} divides the input images into three groups: images of the original class, images of the target class, and original class images that have been successfully attacked. By comparing the active neurons from the three groups and their pathways across neural layers, the authors disclose what alternative pathways were exploited to make the attacks successful.

\begin{figure*}[tbh]
 \centering 
 \includegraphics[width=\textwidth]{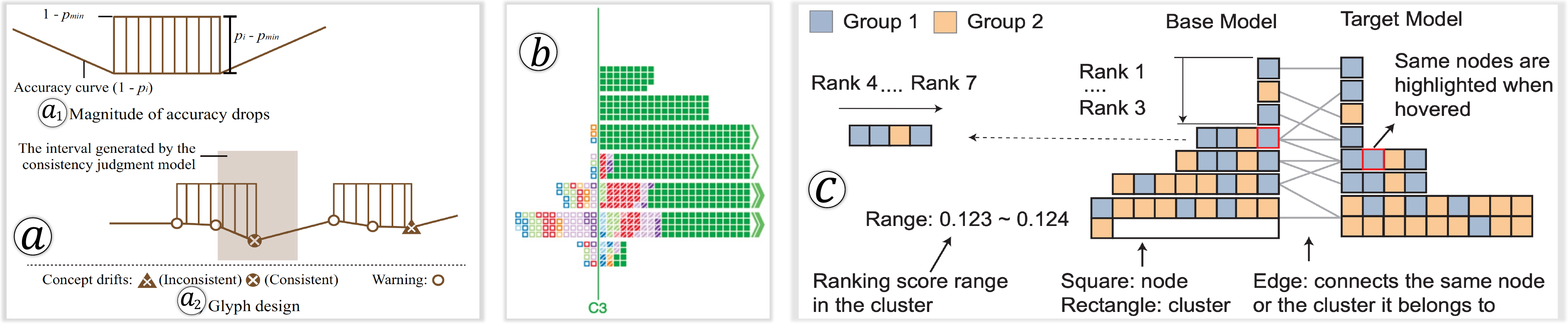}
  \vspace{-0.3in}
 \caption{(a) Glyphs designed to identify concept drift. Image courtesy of Wang et al.~\cite{wang2020conceptexplorer} \copyright \ 2020 IEEE. (b) Each square represents one instance and its vertical position shows the class probability. The square glyphs and their position also encode the prediction correctness. Image courtesy of Ren et al.~\cite{ren2016squares} \copyright \ 2016 IEEE. (c) Graph nodes (in orange and blue) are clustered by their ranking score and nodes of the same cluster are presented in a rectangle for similar exposure. Rankings from two models can also be compared. Image courtesy of Xie et al.~\cite{xie2021fairrankvis} \copyright \ 2021 IEEE.}
 \label{fig:assess}
\end{figure*}

\subsection{Assess Data}
\label{sec:assess}
The VIS4ML efforts in data assessment come from three major directions: (1) monitor the quality of input data to detect data deficiencies; 
(2) assess the output from ML models for their evaluations; (3) diagnose ML models' input and output to disclose biases rooted in both data and models.

\subsubsection{Assess Input Data - Data Quality}
\label{sec:assess_input}
As input data define the performance upper bound of ML models~\cite{zha2023data,strickland2022andrew}, it is crucial to guarantee their quality before training. VIS can help to expose data deficiencies or reveal the drift of data distributions, and thus, has been adopted widely in input data assessment~\cite{wang2020conceptexplorer,yang2020diagnosing}.

ConceptExplorer~\cite{wang2020conceptexplorer} uses a line chart (with glyphs) to monitor the drift level of time series data. Specifically, the sequential data are first fed into a predictive model and concept drifts are detected based on the model's error rate in a sliding time window. The error rate remains stable when there is no drift, but increases abnormally when drift happens. Based on this, the line chart uses strip glyphs to highlight suspicious regions. As shown in Fig.~\ref{fig:assess}($\text{a}_1$), $p_i$ denotes the prediction error at step $i$ and $1{-}p_i$ is the accuracy. $p_{min}$ denotes the minimum error rate in the time window ended at step $i$. The strip glyphs present the magnitude of accuracy drops in the suspicious drift regions. Based on the drop level, different glyphs (e.g., empty circles, filled circles/triangles with a cross) are used to mark important steps in Fig.~\ref{fig:assess}($\text{a}_2$). 
Similarly, DriftVis~\cite{yang2020diagnosing} also monitors the drift level of time-series data with a line chart, in which the drift level is measured through the energy distance between the new-coming and existing data.
OoDAnalyzer~\cite{he2021can} detects out-of-distribution (OoD) samples in test data, whose features are not well-covered by the training data. Superior to conventional methods that only offer an OoD score for a sample, OoDAnalyzer visualizes the sample together with its similar neighbors as a context for investigation. An efficient grid layout algorithm has also been introduced to hierarchically explore enormous data samples and detect the OoD ones.

\subsubsection{Assess Output Data - Performance Analysis}
\label{sec:assess_output}
Evaluating ML models' performance is a fundamental ML task and multiple numerical metrics have been proposed. However, these metrics are often overly aggregated, preventing ML practitioners from gaining performance insights in a finer data granularity. Many novel visualizations have been proposed to address this issue, which visualize models' performance either \textit{after} or \textit{over} their training.

For evaluations \textit{after} model training, Squares~\cite{ren2016squares} is a typical example that improves the confusion matrix visualization for multi-class classifiers. As shown in Fig.~\ref{fig:assess}(b), each square represents a data instance and its vertical position reflects the probability for the corresponding class (i.e., $C3$ here). The squares on the left of the axis (outlined boxes) are $C3$ instances but mis-predicted as other classes (i.e., false negatives). Their color reflects the predicted class. The squares on the right are instances being predicted as $C3$, the solid ones are true positives and the striped ones are false positives (with their color reflecting the true class label). 
For scalability concerns, the squares can be aggregated into strips/stacks and multiple such visualizations can be presented in parallel for multiple classes (Fig.~\ref{fig:instance}($\text{c}$)). 
The design presents not only the confusion matrix but also the prediction confidence, and enables users to interact with individual instances for diagnosis.
Similar examples in this group include Confusion Wheel~\cite{alsallakh2014visual} and ModelTracker~\cite{amershi2015modeltracker}.

The second group of evaluations tracks ML models' performance \textit{over} training to monitor their evolution. For example, Wang et al.~\cite{wang2021investigating} propose a Sankey-diagram based temporal confusion matrix, as we have explained in Fig.~\ref{fig:group}(c). The visualization not only reflects the model's quality, but also tracks the improved and degenerated data instances 
(through the green and red bands between neighboring Sankey nodes) 
for model diagnosis. There are multiple other visualizations revealing the temporal performance evolution for different ML models, e.g.,~\cite{liu2017visual1,hinterreiter2020confusionflow, puhringer2020instanceflow}.

\subsubsection{Assess Fairness - Bias Analysis}
With the rising concerns about fairness in ML, bias analysis becomes increasingly important. Biases can stem from the input data, undesirable trainings (e.g., feature intersections), or the way that data were presented (e.g., content biases).

To study \textit{input data biases}, CoFact~\cite{kaul2021improving} divides input tabular data into three groups based on a feature condition: (1) instances satisfying the condition; (2) instances that do not satisfy the condition but are similar to those in (1) in other features; (3) instances that do not satisfy the condition and are not similar to (1). By comparing the three groups and their feature value distributions, the authors successfully expose the confounding factors in the tabular data. 
For image data, DendroMap~\cite{bertucci2022dendromap} uses treemaps to hierarchically explore a large number of input images. From the exploration, the authors notice that \textit{sunscreen} images often come with lighter skin colors. This feature co-occurrence misleads ML models from learning the right features of \textit{sunscreen}, and should be exposed before model training.

\begin{figure*}[tbh]
 \centering 
 \includegraphics[width=\textwidth]{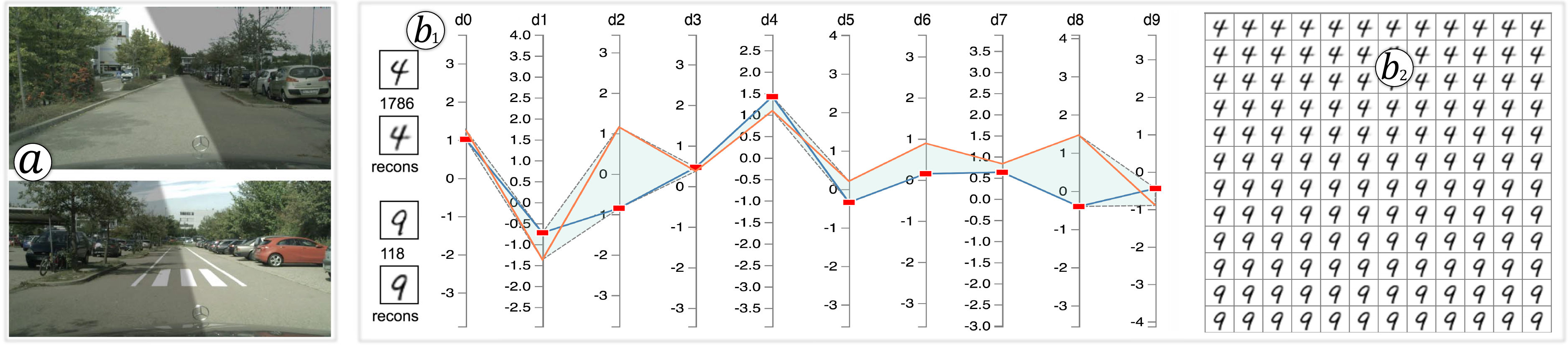}
  \vspace{-0.3in}
 \caption{(a) Images are augmented by adding artificially generated shadows. Image courtesy of Zhao et al.~\cite{zhao2021human} \copyright \ 2021 IEEE. DeepVID generates images between the to-be-interpreted digit `4' and `9' (b2) by interpolating their latent vectors (b1). Image courtesy of Wang et al.~\cite{wang2019deepvid} \copyright \ 2019 IEEE.}
 \vspace{-0.1in}
 \label{fig:generate}
\end{figure*}

To expose \textit{intersectional biases} hidden in well-trained predictive models, FairVis~\cite{cabrera2019fairvis} compares models' performance across feature combinations.
It has been noticed that an ML model with fair performance on individual features may yield unfair performance on feature combinations. For example, a loan eligibility model can generate similar approval rates for \textit{Male} and \textit{Female} applicants, and similar approval rates for \textit{White} and \textit{Black or African American} applicants. However, its approval rates for \textit{Male + White} applicants may be much higher than those of the \textit{Female + Black or African American} applicants. To disclose this, FairVis uses multiple strip plots (Fig.~\ref{fig:comparison}(d)) to compare ML models' performance in subgroups defined by different feature combinations.

\textit{Content biases}, where similar contents were not treated equivalently, have also been examined in VIS4ML. 
For example, graph nodes with similar ranking scores may not be given similar exposures due to their ranking positions. FairRankVis~\cite{xie2021fairrankvis} (Fig.~\ref{fig:assess}(c)) addresses this problem by clustering nodes (squares in blue or orange) based on their ranking scores and organizing nodes of the same cluster into a horizontal rectangle (with black strokes) for equal exposure. In Fig.~\ref{fig:assess}(c), the bottom cluster from the left side has 10 nodes with very similar scores ($0.123{\sim}0.124$) and they are organized into the same rectangle to reduce the content bias that may position them far apart. The system can also compare the rankings from two models (i.e., the ``Base Model'' and ``Target Model'' in the figure).

\subsection{Generate Data}
Data generation extends the dataset $X$ in Eq.~\ref{eq:data} by introducing new instances with desired features. These features can be used to probe ML models' behaviors for better understanding/diagnosis (e.g., ``what-if" analyses) or refine ML models to better cover some corner cases (e.g., adversarial training). This task is very specific to VIS4ML and it is not well-covered in earlier VIS surveys. The essence of data generation is feature augmentation, which can be conducted (1) \textit{directly} in the data space or (2) \textit{indirectly} in a latent space.

\subsubsection{Augment Data Directly in the Data Space}
The features of individual instances are often interpretable, e.g., the \textit{age} and \textit{capital-gain} fields of a tabular census dataset. Their semantics enable users to directly perturb their values and probe ML models' behaviors. For example, the What-If Tool~\cite{wexler2019if} provides a Datapoint Editor View to allow users to directly modify instances' feature values (e.g., increasing the \textit{capital-gain}). By feeding the new instances back to the ML models and checking their performance discrepancy, the users can verify different hypotheses on the models. VIS4ML works based on counterfactual examples, e.g., ~\cite{cheng2020dece,gomez2021advice}, are along the same line and they may rely on automatic algorithms to generate new instances.

Besides tabular data, MD-array data (e.g., images) are also frequently perturbed to probe ML models' behaviors. For example, Bilal et al.~\cite{bilal2017convolutional} generate new image instances by decoloring (i.e., from RGB to gray-scale) or rotating existing ones. By feeding those new images into CNNs, they identify color-invariant and rotation-invariant classes where the CNNs perform well regardless of the corresponding images' color/rotation. Wang et al.~\cite{wang2020hypoml} synthesize two controlled datasets from the original dataset by adding: (1) extra information about a studied concept; (2) random noises that are not related to the concept. The two datasets are then used to train two ML models with the same architecture and configurations, separately. Based on the models' performance discrepancy under the controlled settings, different hypotheses can be verified through statistical significance.

Apart from model understanding and diagnosis, the generated data can also be used to refine ML models. For example, ConceptExtract~\cite{zhao2021human} trains a light-weighted ML model to extract image concepts (e.g., stripe and shadow) learned by a large CNN. Using the system, the users identified a weakness of the CNN in detecting objects with shadows, and overcame it by reinforcing the model to learn more from images with shadows. As shown in Fig.~\ref{fig:generate}(a), more training images are generated by directly adding artificial shadows to the original ones. The CNN fine-tuned on them demonstrated considerable performance improvement.

\subsubsection{Augment Data Indirectly in a Latent Space} 
New data instances can also be generated by encoding the existing instances into a latent space, modifying their latent representations, and decoding them back to the data space. Instances generated in this way often present smooth features with fewer artifacts, as the modifications on their latent representations will impact the reconstructions globally.

For example, DeepVID~\cite{wang2019deepvid} interprets how a CNN differentiates digit `4' and digit `9' images by generating new images smoothly transferring from `4' to `9' to probe the CNN's decision boundary. A VAE encoder is used to transform the two images into a 10D latent space, presented by the PCP in Fig.~\ref{fig:generate}($\text{b}_1$). The orange and blue polylines denote the 10D latent representations of the two images. Then, the two polylines are linearly interpolated inside individual latent dimensions (i.e., within the cyan band). Lastly, by sampling polylines from the interpolated regions and feeding them into the corresponding VAE decoder, semantically meaningful images are generated. As shown in Fig.~\ref{fig:generate}($\text{b}_2$), the generated images present features smoothly transferring from the digit `4' (the top-left one) to `9' (the bottom-right one). Using them, a binary surrogate model can be trained to mimic the original CNN and delineate the decision boundary between the two classes. 

VATLD~\cite{gou2020vatld} and VASS~\cite{he2021can} are similar works along this line, which use  $\beta$VAE and CVAE to extract visual concepts from images and encode them into orthogonal latent dimensions. 
VIS4ML helps to interpret those dimensions and facilitates adversarial training algorithms in manipulating the latent representations. By decoding the new latent representations back to the image space, the authors obtain images with augmented features that can be used to further fine-tune and improve the corresponding ML models.

\subsection{Improve Data}
Refining ML models can be accomplished by optimizing the architectures/hyper-parameters of the models or improving the quality of their input data. As techniques for the former continue to mature, model developers are increasingly recognizing that achieving greater performance gains from the latter is comparatively easier. This results in the rising popularity of data-centric AI~\cite{zha2023data,strickland2022andrew}, recently.
As the data contain two parts, i.e., $X$ and $Y$ in Eq.~\ref{eq:data}, their improvements also come from two aspects, the \textit{features} and \textit{supervision}.

\subsubsection{Improve Features}
The features encoded in individual data instances are what the ML models learn from. Improving them can thus be conducted by curating instances with desired features or selecting/synthesizing better features.

\textit{Instance curation} has been conducted by (1) selecting instances with more desired features, (2) excluding instances with undesired features, and (3) matching the feature coverage in training and test data. For case (1), Ye et al.~\cite{ye2019interactive} introduced an interactive data curation system to guide the training of GANs in generating intended features (e.g., \textit{happy faces}). The system progressively trains multiple binary classifiers to predict if an image includes the contents to be generated or not. These classifiers form a committee to vote out the most disagreed instances, which are then presented to users for manual labeling. 
For case (2), DGMTracker~\cite{liu2017towards} diagnoses deep generative models by disclosing the training details of individual instances, from which, the authors identified training failures caused by outlier instances, e.g., a \textit{plane} image with a large portion of blue sky. They tried to exclude those outliers from training for a quick fix and also proposed theoretical solutions to fix the issue. 
For case (3), as ML models are trained and tested on separate datasets (to avoid over-fitting), ensuring the features of test instances are well covered by the training instances is crucial. For example, a \textit{cat}-\textit{dog} classifier trained on black-\textit{cat} and white-\textit{dog} images will perform badly on a white-\textit{cat} image, which is an OoD sample to the classifier. OoDAnalyzer~\cite{chen2020oodanalyzer} visually identifies such samples from test data through an ensemble OoD detection method and an efficient $k$NN-based grid layout of images. After identifying them, model developers can add the images with the missing features into the training data to fine-tune the ML models. \texttt{Assessing} and \texttt{improving} data happened sequentially in this work.

\textit{Feature selection/synthesis} improves data by adding/excluding/transforming features. The operation differs from instance curation, as it affects all instances rather than some of them. For example, FeatureEnVi~\cite{chatzimparmpas2022featureenvi} helps users generate, transform, and select features to train XGBoost models. The system first ranks the features of tabular data using multiple automatic feature-importance metrics. The rankings then guide users to exclude less important ones. A radial hierarchical graph is introduced to convey the importance of features in different data slices. 
With this graph, the users can decide if a moderately important feature should be excluded or not. The hierarchical graph and embedded glyph visualizations also present statistics (e.g., correlation, mutual information) between features, assisting users in transforming and combining existing features to generate new ones. Similar feature selection and composition works have also been proposed for logistic regressions~\cite{zhang2016visual}, deep sequence models~\cite{ming2019protosteer}, and ensemble models~\cite{wang2022learning,liu2017visual1}.

\subsubsection{Improve Supervision}
Supervision is the annotation information associated with the data that guides the training toward the learning goal. Therefore, clearer and more explicit supervision often leads to easier model training and better model performance.

Training data from various sources often suffer from noisy/missing/incorrect label information. Interactive VIS tools are very effective to incorporate human input and improve the label quality in these cases. 
For example, to correct the mis-labeled training instances, DataDebugger~\cite{xiang2019interactive} proposes a hierarchical layout, enabling users to explore a large number of training samples in a top-down manner. The higher hierarchy levels present fewer samples for an overview, and users can drill down to lower levels for more samples' details. 
This scalable layout provides users an interface to select data instances of interest and they can interactively correct their labels and convert them into trusted instances. An automatic label-error detection algorithm is then applied on them to propagate their labels and further identify other mis-labeled instances for iterative correction.

Despite labels, the supervision can also be other types of annotation. For example, Bilal et al.~\cite{bilal2017convolutional} explore the hierarchy of classes in the ILSVRC 2012 dataset (e.g., both \textit{cat} and \textit{dog} are \textit{mammal}, which is a subclass of \textit{animal}). Integrating the class hierarchy into the training of a CNN, the authors successfully accelerate the training and improve the model's accuracy. 
GenNI~\cite{strobelt2021genni} introduces an interactively defined constraint graph to guide the text-generation process. Following the constraint graph, the ML model first generates/forecasts several output sentences, based on which, the users examine individual outputs and refine the constraint graph. This \textit{Refine-Forecast} paradigm, combining the efforts from both humans and AI, iteratively supervises the model's generative behavior and improves the outputs' quality. There are also works that leverage the information from different modalities of multi-modality data to mutually reinforce the supervision in respective modalities. For example, MutualDetector~\cite{chen2021towards} integrates caption supervision with object detection to improve both the noisy captions and imprecise bounding box information. The work extracts labels from image captions, which are then used to supervise the training of the object detector. The objects extracted from the detector, in return, further improve the captions' quality.

\section{Research Opportunities}
\label{sec:trend}

This section examines the distributions of the \fourthnumber{} papers across the 5 data types, 6 VIS4ML tasks, and their intersections. The distributions reveal which parts of the taxonomy that existing works focus on and which parts have not been sufficiently explored, unveiling potential opportunities.

\begin{figure}[tbh]
 \centering 
 \includegraphics[width=\columnwidth]{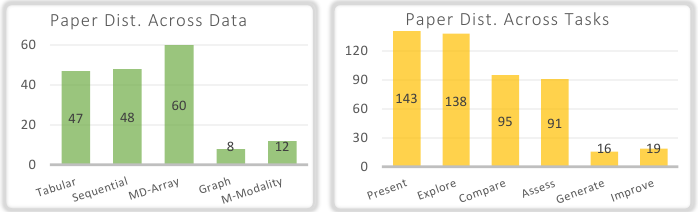}
  \vspace{-0.28in}
 \caption{Paper distributions across 5 data types and 6 VIS4ML tasks.}
 \label{fig:dist}
\end{figure}

\subsection{Opportunities From Data Types}
From the data type distribution (Fig.~\ref{fig:dist}, left), it is very obvious that existing VIS4ML works focus more on \texttt{tabular}, \texttt{sequential}, and \texttt{MD-array} data, whereas the \texttt{graph} and \texttt{multi-modality} data are less covered.

\textit{Opportunity 1: interpreting ML models for graph data.} Compared to the first three data types, \texttt{graph} data is more irregular and difficult to handle, especially heterogeneous graphs. However, we envision more VIS4ML works will come for this type of data for the following reasons. First, graph is a powerful way to structurally 
organize data and convey their relational information, e.g., a citation graph connects discrete papers and builds relationships among them. Its unique merits keep the amount of graph data consistently growing. Second, advanced graph learning models, e.g., GNNs, are also evolving fast, so as to the demanding need for their understanding, diagnosis, and refinement.

\textit{Opportunity 2: coordinated analysis of multiple data modalities with multiple ML models.} 
We have observed an increasing number of ML works that integrate the learning outcomes from different modalities of \texttt{multi-modality} data for better performance. For example, the sentiment analysis model in~\cite{wang2021m2lens} is trained on facial expressions (video), voices of the speakers (audio), and the corresponding textual transcripts (text). Multiple ML models are often involved in these works to take their respective advantages in handling different data modalities. Given the popularity of these ML works, two VIS directions are very promising for this data type. \textit{First, exploring and relating different modalities of multi-modality data with coordinated multiple views.} Coordinated visual explorations have been repetitively verified to be effective in handling multi-faceted data~\cite{wang2018visualization}, and the techniques are readily transferable to the increasingly complex multi-modality data from ML. 
\textit{Second, mutual enhancement of the information between different data modalities.} The underlying connections between different data modalities can be used to mutually reinforce the information inside each. 
This is a good way to \texttt{improve} data by leveraging the implicit information inside a modality as explicit supervision for the other.
For example, MultualDetector~\cite{chen2021towards} improves the noisy image captions and imprecise bounding boxes of image objects by borrowing the information from each other as supervisions. VIS plays a critical role here, as it provides the necessary guidance to better bridge different data modalities and facilitates the mutual enhancement between them.

\subsection{Opportunities From Data-Centric Tasks.} 
From the paper distribution over the six tasks (Fig.~\ref{fig:dist}, right), \texttt{presenting} and \texttt{exploring} data are the most fundamental tasks conducted by most VIS4ML works. Only a few short papers with static visualizations do not involve data exploration.
\texttt{Comparing} and \texttt{assessing} data are also commonly performed for model understanding/diagnosis. However, fewer works cover the tasks of \texttt{generating} and \texttt{improving} data. These two majorly contribute to model refinements but have not been sufficiently explored.

\textit{Opportunity 3: model refinement with data generation and improvement.} With the rapid evolution of XAI, researchers are no longer satisfied with works that only help to understand or diagnose ML models, but are eager to see how VIS can help to further refine the corresponding models. In practice, model \textit{understanding} and \textit{diagnosis} are often the prerequisites for model \textit{refinement}. In the era of data-centric AI~\cite{strickland2022andrew}, we believe many model refinement opportunities reside in data \texttt{generation} and \texttt{improvement}. These two tasks generate data with desired features or improved supervisions to refine ML models from the data perspective. They help to convert insights obtained from model \textit{understanding} and/or model \textit{diagnosis} into direct model \textit{refinement} actions, demonstrating the very practical role that VIS can play.

\textit{Opportunity 4: more smartly involving humans into the data-centric VIS4ML tasks, but minimizing their labor effort.}
For all six tasks, especially the last two, the inputs from humans often play important roles, e.g., label corrections with users' prior knowledge on different classes~\cite{xiang2019interactive}. 
In fact, human-in-the-loop analyses have been adopted in many VIS4ML works~\cite{zhao2021human, ye2019interactive}.
Nevertheless, some works still require intensive human interventions, making the explorations or analyses not friendly enough to users. 
Therefore, it is worth more efforts to better team up humans and AI to leverage humans' intelligence but minimize their labor effort in the meantime. Some seminal VIS4ML works, e.g.,~\cite{gou2020vatld, strobelt2021genni}, have started this kind of explorations, e.g., asking humans to provide key controls only and leaving the heavy-lifting part to automatic AI algorithms.  
This direction also opens up the opportunity to more effectively combine human-computer interaction (HCI) and VIS techniques to better serve ML.

\subsection{Opportunities From Data-Task Intersections}
\begin{figure}[tb]
 \centering 
 \includegraphics[width=\columnwidth]{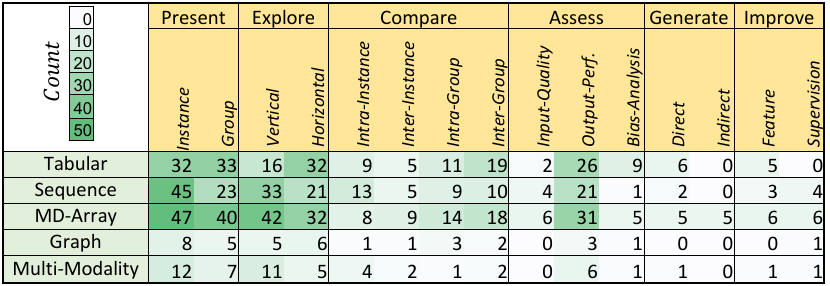}
  \vspace{-0.3in}
 \caption{Joint distribution of papers across data and task sub-categories.}
 \vspace{-0.15in}
 \label{fig:crossdist}
\end{figure}

Fig.~\ref{fig:crossdist} shows the cross-distribution between the five data types and six tasks. 
From it, more papers are distributed on the top-left, i.e., the intersections between the first three data types and the first four tasks, echoing the two marginalized distributions in Fig.~\ref{fig:dist}. From the lighter color cells, we have identified several more research opportunities.

\textit{Opportunity 5: input data assessment and bias analysis for data-centric AI.} The \texttt{output-performance} analysis currently dominates the \texttt{assess} task in Fig.~\ref{fig:crossdist}. With more efforts on data-centric AI, we look forward to the increase of \texttt{input-quality} assessment works and they will cover more diverse data types (e.g., the graph and multi-modality data). Also, the \texttt{input-quality} assessment is the prerequisite for further data \texttt{improvement} and/or \texttt{generation}, echoing 
 the earlier \textit{Opportunity 3}.
Moreover, with the rising concerns on model fairness, we would also expect the number of \texttt{bias-analysis} works from the \texttt{assess} task to increase. Existing works in this category mostly focus on \texttt{tabular} data, as the semantically meaningful tabular features (e.g., \textit{gender} and \textit{race}) could be naturally considered as protected features. However, biases do exist in other data types, e.g., undesired feature co-occurrences in \texttt{MD-array} data~\cite{bertucci2022dendromap} or unfair node exposures in \texttt{graphs}~\cite{xie2021fairrankvis}, and more works are waiting to be proposed to fill this gap.

\textit{Opportunity 6: more general and scalable visualizations for heterogeneous and large-scale data.} From Fig.~\ref{fig:crossdist}, we also find that some data-centric tasks have only been performed on one data type due to the limited generalizability of the corresponding tasks (or VIS techniques). For example, the \texttt{indirect} data \texttt{generation} has only been explored on \texttt{MD-array} data, mostly images. This is largely due to the success of CNN-based encoder-decoder frameworks. On the other hand, it also indicates a great research opportunity in data generation for other data types using the \texttt{indirect} manner. To better take care of the data heterogeneity, a general solution that can accomplish a data-centric task across all different data types is very preferable. Furthermore, as ML models are often trained on a large number of input data instances and generate a massive amount of intermediate (e.g., activations from DNNs) and output data, extending existing visualizations to make them more scalable is also a promising direction. 
For example, DendroMap~\cite{bertucci2022dendromap} explores large-scale image datasets through hierarchically clustering the HD image representations and enabling users to explore their interested images via interactions. A better understanding of the images leads to a better comprehension of the corresponding ML models' behavior.

\definecolor{tabrule}{RGB}{220,220,220}%
\definecolor{tabred}{RGB}{230,36,0}%
\definecolor{tabgreen}{RGB}{0,116,21}%
\definecolor{taborange}{RGB}{255,124,0}%
\definecolor{tabbrown}{RGB}{171,70,0}%
\definecolor{tabyellow}{RGB}{255,253,169}%
\definecolor{tabrule}{RGB}{220,220,220}%
\definecolor{tabdata}{RGB}{132,172,78}
\definecolor{tabmod}{RGB}{252,224,118}

\begin{table}[]\tiny
\caption{All \dlnumber{} papers here (sorted by time and venue in \texttt{ReasonCode.xlsx}) are \textbf{DL-related}. (Abbreviations: G: \textit{Group}; I: \textit{Instance}; H: \textit{Horizontal}; V: \textit{Vertical}; IaI: \textit{Intra-Instance}; IeI: \textit{Inter-Instance}; IaG: \textit{Intra-Group}; IeG: \textit{Inter-Group}; IQ: \textit{Input-Quality}; OP: \textit{Output-Performance}; BA: \textit{Bias-Analysis}; D: \textit{Direct}; I: \textit{Indirect}; F: \textit{Feature}; S: \textit{Supervision}). }
\taburulecolor{tabrule}
\label{tbl:sum}
\centering
\renewcommand{\arraystretch}{1.167} 
\setlength\tabcolsep{5pt} 
\begin{tabular}{r|ccccc|cccccc|}
\cline{2-12}
\multicolumn{1}{l!{\color{black}\vrule}}{}   & \multicolumn{5}{c!{\color{black}\vrule}}{\cellcolor{tabdata}{Data Type}}                                                                                                     & \multicolumn{6}{c!{\color{black}\vrule}}{\cellcolor{tabmod}{Data-Centric VIS4ML Tasks}}                                                                                                                                  \\ \cline{2-12} 
\multicolumn{1}{r@{}!{\color{black}\vrule}}{Publication}             & \multicolumn{1}{p{0.001cm}|}{\rotatebox{90}{Tabular}} & \multicolumn{1}{p{0.001cm}|}{\rotatebox{90}{Sequence}} & \multicolumn{1}{p{0.001cm}|}{\rotatebox{90}{MD-Array}} & \multicolumn{1}{p{0.001cm}|}{\rotatebox{90}{Graph}} & \multicolumn{1}{p{0.001cm}!{\color{black}\vrule}}{\rotatebox{90}{Hybrid}} & \multicolumn{1}{c|}{\rotatebox{90}{Present}} & \multicolumn{1}{c|}{\rotatebox{90}{Explore}} & \multicolumn{1}{c|}{\rotatebox{90}{Compare}} & \multicolumn{1}{c|}{\rotatebox{90}{Assess}} & \multicolumn{1}{c|}{\rotatebox{90}{Generate}} & \multicolumn{1}{c!{\color{black}\vrule}}{\rotatebox{90}{Improve}} \\ \hline

\rowcolor[HTML]{F5F5F5}\multicolumn{1}{r@{}!{\color{black}\vrule}}{Liu et al.~\cite{liu2016towards}} &\multicolumn{1}{c|}{} &\multicolumn{1}{c|}{} &\multicolumn{1}{c|}{\cellcolor{tabdata}} &\multicolumn{1}{c|}{} &\multicolumn{1}{c!{\color{black}\vrule}}{} &\multicolumn{1}{c|}{ \cellcolor{tabmod}G}&\multicolumn{1}{c|}{ \cellcolor{tabmod}H}&\multicolumn{1}{c|}{ }&\multicolumn{1}{c|}{ \cellcolor{tabmod}OP}&\multicolumn{1}{c|}{ }&\multicolumn{1}{c!{\color{black}\vrule}}{ }\\
\hline

\end{tabular}
\end{table}

\begin{table}[htb]\tiny
\caption{Tab.~\ref{tbl:sum} continued. All \mlnumber{} papers in this table are for \textbf{classic ML models}.}
\vspace{-0.1in}
\taburulecolor{tabrule}
\label{tbl:sum2}
\centering
\renewcommand{\arraystretch}{1.15} 
\setlength\tabcolsep{5pt} 
\begin{tabular}{r|ccccc|cccccc|}
\cline{2-12}
\multicolumn{1}{l!{\color{black}\vrule}}{}   & \multicolumn{5}{c!{\color{black}\vrule}}{\cellcolor{tabdata}{Data Type}}                                                                                                     & \multicolumn{6}{c!{\color{black}\vrule}}{\cellcolor{tabmod}{Data-Centric VIS4ML Tasks}}                                                                                                                                  \\ \cline{2-12} 
\multicolumn{1}{r@{}!{\color{black}\vrule}}{Publication}             & \multicolumn{1}{p{0.001cm}|}{\rotatebox{90}{Tabular}} & \multicolumn{1}{p{0.001cm}|}{\rotatebox{90}{Sequence}} & \multicolumn{1}{p{0.001cm}|}{\rotatebox{90}{MD-Array}} & \multicolumn{1}{p{0.001cm}|}{\rotatebox{90}{Graph}} & \multicolumn{1}{p{0.001cm}!{\color{black}\vrule}}{\rotatebox{90}{Hybrid}} & \multicolumn{1}{c|}{\rotatebox{90}{Present}} & \multicolumn{1}{c|}{\rotatebox{90}{Explore}} & \multicolumn{1}{c|}{\rotatebox{90}{Compare}} & \multicolumn{1}{c|}{\rotatebox{90}{Assess}} & \multicolumn{1}{c|}{\rotatebox{90}{Generate}} & \multicolumn{1}{c!{\color{black}\vrule}}{\rotatebox{90}{Improve}} \\ \hline

\rowcolor[HTML]{F5F5F5}\multicolumn{1}{r@{}!{\color{black}\vrule}}{Ren et al.~\cite{ren2016squares}} &\multicolumn{1}{c|}{} &\multicolumn{1}{c|}{} &\multicolumn{1}{c|}{\cellcolor{tabdata}} &\multicolumn{1}{c|}{} &\multicolumn{1}{c!{\color{black}\vrule}}{} &\multicolumn{1}{c|}{ \cellcolor{tabmod}I, G}&\multicolumn{1}{c|}{ \cellcolor{tabmod}V}&\multicolumn{1}{c|}{ }&\multicolumn{1}{c|}{ \cellcolor{tabmod}OP}&\multicolumn{1}{c|}{ }&\multicolumn{1}{c!{\color{black}\vrule}}{ }\\
\hline

\end{tabular}
\end{table}

\section{Discussion and Limitations}
\label{sec:discussion}

Our survey has several inherent limitations. \textit{First}, our taxonomy is inevitably impacted by our view of the VIS4ML problem. Although the co-authors all have years of experience working on ML and VIS, certain choices of the papers and categorizations have been influenced by our past experience. This limitation inherits from the subjective nature of a survey paper. However, as our taxonomy well-covers the majority of the VIS4ML literature and our analysis comes with concrete statistics, we are confident to believe that our survey provides valuable insights into this area.

\textit{Second}, there are also subjective decisions over the coding of individual papers. For example, some papers focused on proposing solutions to \texttt{compare} ML models, but also presented brief cases that slightly \texttt{improve} the models. Whether coding the papers with the \texttt{improve} task or not is thus subjective. To mitigate this problem, we provide a spreadsheet in our Supplementary Material, i.e., \texttt{ReasonCode.xlsx}, summarizing all the \fourthnumber{} papers and briefly explaining why we code individual papers into their respective categories. Readers can use the spreadsheet to understand our coding rationales and suggest different codings. We believe the well-documented reasons will help to track and improve our labeling of the papers. 

\textit{Lastly}, there are many other venues with VIS4ML works (e.g., CHI, IUI, and ACL) that we could not conduct an exhaustive search on, due to the limited length of this survey. We have considered selectively including some papers from them as they are equivalently important. However, it would involve more subjective decisions and make the survey less self-contained. Furthermore, those venues also have their respective focuses beyond VIS (e.g., ML or HCI). In contrast, the current six venues we have covered all have a dominant focus on VIS. Considering these factors, we made the deliberate choice to confine our survey within the six VIS venues only, rather than inundating readers with VIS4ML contributions from a diverse array of sources. We hope our survey can work as a starting point to pique readers' interest in reexamining VIS4ML papers, even for those outside of the six venues, through a data-centric lens.

\section{Conclusion}
\label{sec:conclusion}

In this paper, we review the latest VIS4ML works (\fourthnumber{} papers) from the past seven years and introduce a data-centric taxonomy to organize them. Our taxonomy first identifies the data types that individual works have focused on and categorizes them into five groups. Then, focusing on the VIS operations applied to these data, we elicit six data-centric VIS4ML tasks and explain how individual tasks have been conducted.
Lastly, based on our review and the paper distribution, we provide insights into the current VIS4ML endeavors and envision future research directions.   

\ifCLASSOPTIONcaptionsoff
  \newpage
\fi



\bibliographystyle{IEEEtran}
\bibliography{IEEEabrv,./tvcg19}
%



%

\vspace{-0.1in}
\begin{IEEEbiographynophoto}{Junpeng Wang}
is a research scientist at Visa Research. He received his B.E. degree in software engineering from Nankai University, M.S. degree in computer science from Virginia Tech, and Ph.D. degree in computer science from the Ohio State University. His research interests are broadly in visualization, visual analytics, and explainable AI.
\end{IEEEbiographynophoto}

\vspace{-0.2in}
\begin{IEEEbiographynophoto}{Shixia Liu}
a professor at Tsinghua University. Her research interests include visual text analytics, visual social analytics, interactive machine learning, and text mining. She worked as a research staff member at IBM China Research Lab and a lead researcher at Microsoft Research Asia. She received a B.S. and M.S. from Harbin Institute of Technology, a Ph.D. from Tsinghua
University. She is a fellow of IEEE and an associate editor-in-chief of IEEE Trans. Vis. Comput. Graph.
\end{IEEEbiographynophoto}

\vspace{-0.2in}
\begin{IEEEbiographynophoto}{Wei Zhang}
is a principal research scientist and research manager at Visa Research and interested in big data modeling and advanced machine learning technologies for payment industry. Prior to joining Visa Research, Wei worked as a Research Scientist in Facebook, R\&D manager in Nuance Communications and also worked in IBM research over 10 years. Wei received his Bachelor and Master degrees from Department of Computer Science, Tsinghua University.
\end{IEEEbiographynophoto}




\end{document}